\documentclass[journal]{IEEEtran}

\usepackage{tabularx}
\usepackage{placeins}
\usepackage{ifpdf}
\usepackage{tikz}
\usetikzlibrary{shapes.symbols,decorations.pathreplacing}
\usepackage{xcolor}

\definecolor{avteal}{HTML}{16736E}
\definecolor{avgreen}{HTML}{4C9A2A}
\definecolor{avnavy}{HTML}{1F3864}
\definecolor{avorange}{HTML}{E07B16}
\definecolor{avred}{HTML}{B3251E}
\usepackage{cite}
\usepackage{booktabs}
\usepackage{graphicx}
\ifCLASSINFOpdf
    \usepackage{graphicx}
\else
    \usepackage{graphicx}
\fi
\usepackage[colorlinks=true,
            linkcolor=blue,
            citecolor=blue,
            urlcolor=blue]{hyperref}
\begin{document}
\title{Quantifying the Impact of Ambulance Ramping: A Multi-Year Analysis of Victorian Emergency Medical Services Cases}
\author{Ayesha~Tanveer,
        Khandakar~Ahmed,
        Assefa~Teshome,
        Oyetunde~Gbadeyan,
        Ziad~Nehme
        % <-this % stops a space
\thanks{Ayesha Tanveer, Khandakar Ahmed and Assefa Teshome are with the Institute for Sustainable Industries and Liveable Cities (ISILC), The Centre of Excellence in Paramedicine (CoEP), Victoria University, Melbourne, Australia, e-mail: (ayesha.tanveer@vu.edu.au).}% <-this % stops a space
\thanks{Ziad Nehme and Oyetunde Gbadeyan are with Center for Research and  Evaluation Ambulance Victoria, Blackburn North, Victoria, Australia.}% <-this % stops a space
}
%\thanks{Manuscript received ; revised .}}
\markboth{September~2026}%
{Shell \MakeLowercase{\textit{et al.}}: Bare Demo of IEEEtran.cls for IEEE Journals}
\maketitle
\begin{abstract}
Ambulance ramping, the delay between hospital arrival and patient handover, is a critical operational bottleneck in Emergency Medical Services (EMS), yet its systemic magnitude and dynamics remain inadequately characterised at scale. This paper quantifies the scale, trajectory, and operational correlates of ramping across an entire statewide EMS system, analysing 2,850,575 ambulance attendances in Victoria, Australia from January 2020 to March 2024 using an Exploratory Data Analysis (EDA). After systematic preprocessing, an analytical cohort of 2,026,569 Emergency Department (ED) transports across 59 hospitals with ED and 79 Local Government Areas (LGA) was examined through interval decomposition, Pareto concentration, hourly cross-correlation, hospital arrival concurrency and priority-stratified operational comparisons. 
Cumulative Ambulance Hours Lost (AHL) totalled 1,491,127 hours, equivalent to approximately 96 ten-hour ambulance shift lost every day of the study window. Ten of 59 hospitals account for 57.8\% of lost hours from 50.9\% of cases. Annual losses rose 57\% to a 2022 peak while transported demand fell 3.7\%, indicating deterioration in per-case handover rather than growth in demand. Handover duration varies little with patient acuity, but rises monotonically with the number of ambulances arriving at the same hospital in the preceding hour, an effect persisting within every hour of the day. Hourly demand is moderately associated with ramping two to four hours later (r = 0.365). These findings establish the empirical preconditions for hospital-state aware ambulance routing.

\end{abstract}
% \fi
\IEEEpeerreviewmaketitle

\section{Introduction}
Emergency Medical Services (EMS) worldwide face sustained growth in demand driven by demographic change, rising population, comorbidity, and shifting healthcare use \cite{Hussain2023AmbulanceDelays, Andrew2020Drivers, CEKLIC2022104886, Dinh01112016}. Studies show increasing ambulance case loads and longer response times, raising concerns about system capacity and service quality and equity \cite{nehme2016factors, jensen2024racing, https://doi.org/10.1002/ams2.70017}. In Denmark, emergency ambulance dispatches rose 43\% over a decade, with longer response times across all priority levels, indicating systemic strain on prehospital care \cite{jensen2024racing}. In Victoria, Australia, EMS demand has grown by about 4.3\% annually (pre-COVID)\cite{hemingway2026new} and statewide emergency road incidents rising from 633,960 in 2019-20 to 722,894 in 2024-25 (an increase of approximately 14.0\%) \cite{AmbulanceVictoria2024AnnualReport}, with evidence that rising case loads now negatively affect response times for time-critical patients \cite{https://doi.org/10.1111/1742-6723.13181}. Comparable trends internationally underscore the need to understand how operational bottlenecks propagate through EMS systems \cite{https://doi.org/10.1002/ams2.70017}.\\One of the contributors to EMS inefficiency is ambulance ramping, defined as the delay between hospital arrival and patient handover in the Emergency Department (ED). While ramping is widely recognised as a manifestation of ED crowding \cite{kingswell2017concepts}, its effects extend beyond hospital boundaries into upstream EMS operations. Empirical studies have shown that offload delays reduce ambulance availability and are associated with prolonged response times for subsequent calls, effectively linking hospital congestion to prehospital performance degradation \cite{jensen2024racing, Dyson2022AmbulanceDelays, li2026empirical, 10.1001/jamanetworkopen.2024.51022}.\\Ramping is highly variable across hospitals, regions, times of day, and demand levels, and hospital accessibility is therefore a dynamic phenomenon rather than a static geographic problem: routing decisions should account for temporally evolving hospital congestion rather than relying solely on travel distance or historical averages \cite{Morley2018EDCrowding, BIMPOU2020102892}.
While studying ramping, there emerge a few concerns. One of those concerns is about where the delay accumulate. EMS operations involve several consecutive stages, including call receiving, ambulance assignment and dispatch, care at the scene, transporting the patient, handover to the ED/transfer of care, and clearing the ambulance, and delays may occur at various steps throughout this process. Existing ambulance offload delay research also uses heterogeneous time definitions and measurement points, ranging from triage to clinical location intervals, to hospital arrival to transfer of care intervals, making it difficult to characterise how delay accumulates across the broader EMS workflow. Dispatch systems further prioritise incidents according to patient acuity, meaning resource constraints may have unequal effects across acuity levels. Although ED crowding and ambulance ramping have been extensively associated with increased delays \cite{VANDERLINDEN201627, wickman2017effect, li2018review} and reduced operational performance, little attention have been given to how bottlenecks such as redistribute delays across the entire EMS workflow, and whether this distribution varies systematically by patient acuity\cite{li2018review}.
Another concern is about the analytical approach used. Existing EMS studies commonly begin by characterising temporal demand, response time distributions, hospital variation and spatial demand hotspots before proposing forecasting or resource allocation strategies. Jensen et al. \cite{jensen2024racing} analysed long-term trends in ambulance demand and response times to identify emerging system pressures; Neira et al. \cite{neira2025novel} employed spatiotemporal descriptive analysis to identify demand patterns before integrating these insights into a predictive model. Although such exploratory analyses are fundamental to understanding EMS dynamics, they are generally presented as preliminary description rather than as structured analytical frameworks in their own right.
Exploratory Data Analysis (EDA) \cite{tukey1977exploratory} is widely recognised as a foundational step in the data science and machine learning pipeline involving statistical summaries, visualisation techniques, and data profiling methods to understand the underlying structure, relationships, and quality of a dataset \cite{Carranza2023}. Recent work recognises EDA as a critical analytical phase that informs subsequent modelling decisions, facilitates the detection of hidden patterns and anomalies, validates underlying assumptions and reduces the risk of model mis-specifications \cite{KAHAGALAGE2024123150, dhummad2025imperative}. Its role is particularly important in EMS research owing to the high dimensionality, heterogeneity, and operational complexity of ambulance data. Large EMS datasets, including the National Emergency Medical Service Information System (NEMSIS), require substantial preliminary exploration to identify viable research objectives, as the dataset’s structure, feature variation, and completeness strongly shape research questions and analytic approaches \cite{williams2022considering}. Moreover, studies assessing EMS data quality have documented considerable missing data and heterogeneity in several core operational variables \cite{abir2021data}. Studies of ambulance response times and operational workflows have used exploratory techniques to analyse response interval distributions, geographic disparities, and environmental influences on service delivery \cite{hill2025understanding}, while studies analysing dispatch-to-availability intervals have applied descriptive statistics to uncover bottlenecks in ambulance utilisation and patient transport \cite{farhat2024exploring}. Research on data harmonisation across ambulance services in Australia and New Zealand highlights significant heterogeneity in variable definitions, collection practices, and schema, which limits comparability and complicates downstream modelling \cite{bunting2023mastering}, while broader public health applications, analyses of violence, drug related incidents, and frequent utilisation patterns, rely on spatial clustering, temporal trend analysis, and descriptive statistics to identify high risk populations and geographic hotspots \cite{hibdon2024use}.
Treated as a primary analytical framework rather than a preliminary step, EDA serves three functions. In \emph{problem formulation}: real-world problem definitions evolve through interaction with the data using techniques such as correlation analysis, distribution visualisation, and dimensionality reduction to identify whether meaningful class separability exists or whether continuous relationships dominate, thereby supporting hypothesis generation and problem definition \cite{williams2022considering, ehlers2023description, chuah2025edamame, zhang2024exploratory, Smith-Miles2011}. In \emph{model selection}: by exposing feature distributions, linearity, multicollinearity, sparsity, and outliers, EDA guides the choice of modelling technique, ensuring only relevant variables are used \cite{fu2024exploratory}, informing the trade-off between interpretable and black-box models \cite{chuah2025edamame}, and matching patterns such as temporal trends, missingness, and spatial decay to appropriate deep learning, statistical, or NLP approaches \cite{abir2021data}. In \emph{model validation}: data quality issues identified during EDA directly shape evaluation strategies and metric selection, supporting robust and unbiased assessment \cite{abir2021data}. Despite these potentials, the application of EDA in EMS research remains largely confined to inspection preceding regression or predictive modelling, and its use as a primary analytical framework for examining system behaviour remains limited. 
By analysing a large-scale EMS dataset across temporal, spatial, and operational dimensions, within a structured exploratory framework adapted from the general EDA framework \cite{de2007exploratory}, this study aims to clarify how ambulance ramping propagates through the EMS system and to identify points of intervention beyond prehospital resource allocation.
This study extends the work \cite{cone2012analysis, vandeventer2011association, hitchcock2010effects,mallows2025ambulance} on ambulance offload delay or ramping by deconstructing the ambulance service interval into its constituent operational phases and stratifying analysis by patient acuity. Total ambulance service time is deconstructed into five operational components, and delays are examined according to the patients' acuity/dispatch categories (Priority 0, Code 1-3) and their transport priorities (1-4) of \cite{AmbulanceVictoria2019AnnualReport, AmbulanceVictoria2024AnnualReport} \footnote{Priority 0, a subset of Code 1, indicates the most urgent events requiring a time-critical response, usually involving patients with life-threatening conditions. Code 1 incidents require urgent paramedic and hospital care. Code 2 incidents are acute and time sensitive, but do not require a lights and sirens response. Code 3 incidents are not urgent but still require an ambulance response \cite{AmbulanceVictoria2024AnnualReport, Nehme2016Timeliness}}).

\section{Methodology }  This section describes the study design, data source, operational definitions, and analytical framework used to explore ambulance ramping in Victoria, Australia. 
\subsection{Study Design \& Setting } 
\begin{figure}[!t]
\centering
\includegraphics[width=\columnwidth]{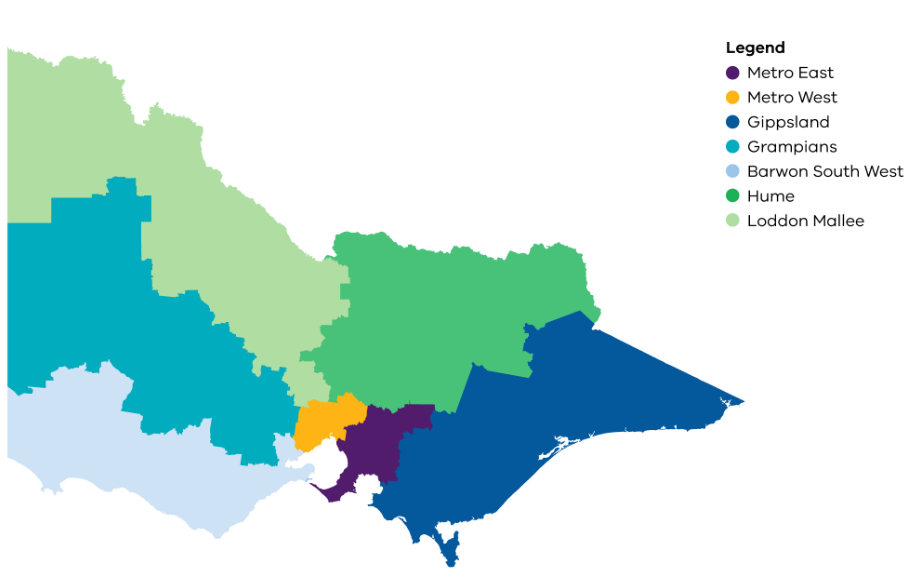}
\caption{Seven AV regions: AV's services are divided into seven regions across the state : the metropolitan (East,West) and five regional and rural regions (Barwon South West, Gippsland, Grampians, Hume and Loddon Mallee). }
\label{fig:AV_Regions}
\end{figure}

This retrospective cohort study included ambulance attendances in Victoria, Australia from 01/01/2020 to 31/03/2024. Ethical approval was obtained from the Victoria University Human Research Ethics Committee (HRE25-105) and Ambulance Victoria’s (AV) Research Committee (R25-023). AV is the sole statutory ambulance provider in Victoria, delivering prehospital emergency care and emergency and non-emergency transport across metropolitan, regional, and rural areas, serving over 7 million people in 2025 \cite{ABS2026PopulationDec2025} across more than 227,000 square km (Fig. \ref{fig:AV_Regions}, adapted from \cite{VAHI_AmbulanceCare_2019}). Emergency responses are managed via the Triple Zero Victoria \cite{TripleZeroVictoria} communications system, which dispatches ambulance resources statewide for patient assessment, prehospital treatment, transport, and retrieval. Trained emergency medical dispatchers triage calls and assign resources using a dispatch matrix\cite{AmbulanceVictoria2019AnnualReport} \cite{VAGO2014EmergencyICT}. %all Triple Zero (000) ambulance transports in Victoria, Australia, 
\subsection{Data Source }A retrospective study of 2,850,575 (2.85 million) emergency ambulance cases attended by AV paramedics from January 2020 to March 2024 was undertaken to investigate ambulance offload delay across the study period. 
\subsection{Study Cohort} The dataset comprised 2,850,575 ambulance dispatches attended in Victoria between January 2020 and March 2024, a window of 1,553 days. Each record represents a single case and carries the timestamps marking each stage of the ambulance operational cycle, together with the dispatch and transport priority codes assigned to it, the destination facility, and the Local Government Area in which the incident occurred.
Destinations were recorded across 283 distinct medical facilities. Of these, 59 were ED capable, the remaining were non-ED facilities including subacute care, residential aged care and specialist ambulatory services, and interstate destinations.
Incidents were recorded across 100 distinct Local Government Areas (LGA). 79 Victorian LGAs are mapped to the 6 AV regions shown in Fig. \ref{fig:AV_Regions}. We have used 'Metro' both for 'Metro East' and 'Metro West' in the paper. The remaining 21, accounting for 11,330 cases (0.42\%), comprised interstate border councils in New South Wales, South Australia and Tasmania together with Unincorporated Victoria.
\subsection{Operational Definitions}
This section defines the key operational terms and time intervals used throughout the study. Each interval is defined with reference to specific timestamp fields recorded in the dataset, ensuring that the definitions are reproducible from the source data.\\ 
\textit{Ambulance Ramping}: Ramping, a term originating from Australian EMS operational reporting \cite{AmbulanceVictoria2019AnnualReport}, refers to the delay experienced by ambulance crews, when a patient cannot be transferred to hospital care due to hospital capacity constraints, after arrival at a hospital. In this study, ramping duration is operationalised as the interval between the ambulance arriving at the destination hospital and the patient being offloaded from the ambulance stretcher measured in minutes.\\
\textit{Ambulance Hours Lost (AHL)}: A composite operational metric used in this study to quantify the cumulative operational cost of delayed transfer of care/ramping at the system, regional, and hospital levels \cite{SAHealthAmbulanceWaitingTimes, AACE2018AnnualReport}. Building on this established concept, we adopted AHL as the cumulative ramping time for framing: total ramping hours across all cases for system level burden quantification; and peak concurrent ramping instances for operational stress measurement.\\ 
\textit{Ambulance shift}: An ambulance shift refers to a rostered period of duty assigned to an EMS employee, with varying durations of up to 14 hours and specified start and finish times. For the purposes of this study, an ambulance shift is operationally defined as a 10-hour period of duty to provide a consistent time window for the analysis \cite{ambulance_victoria_enterprise_agreement_2024}.\\
\textit{Dispatch Code}: The dispatch code (Code 1 = most urgent, requiring a lights and sirens response, to Code 3 = least urgent) is determined by the patient's assessment at the time of 000 call; Priority 0 ( = immediately life-threatening) is included within Dispatch Code 1 \cite{VAHI2024AmbulanceDataset}.\\
%The priority code is assigned by the Triple Zero (000) operator at call receipt on a scale from 0 (most urgent) to 9 (least urgent). 
\textit{Transport Priority}: The level of transport urgency given to a patient according to their clinical status and need for care at the time of ambulance transfer. A patient's condition can change between the initial Triple Zero (000) call and the later assessment by paramedics. Transport Priority 1 indicates the greatest urgency, while Priority 4 indicates the least.\\
\textit{Medical facility with an ED}: A hospital classified as having ED capability, identified through the reference table constructed with the help of AV for this study\cite{VAHI2024AmbulanceDataset}. All ramping analyses restrict the cohort to cases transported to a medical facility with an ED.\\
\textit{Dispatch to scene Time}: The interval between ambulance dispatch and arrival at the scene, reflecting the time from operational commitment of a resource to on-scene availability.
\textit{On Scene Time}: The interval between arrival at the scene and departure from the scene with the patient (only if transported). This interval reflects the time required for on scene clinical assessment, treatment initiation, and patient loading.\\
\textit{Transport Time}: The interval between scene departure and arrival at the destination hospital, reflecting the ambulance travel duration from scene to hospital.\\
\textit{Clearing Time}: The interval between patient offload and ambulance clearance, reflecting post handover activities such as documentation, restocking, and preparation for the next case.\\
% \textit{Emergency and Non Emergency Classification}: Cases are classified as either emergency or non-emergency reflects the operator's initial assessment of call urgency and is set before the ambulance is dispatched to the event location LGA.\\
\begin{figure}[!t]
\centering
\includegraphics[width=\columnwidth]{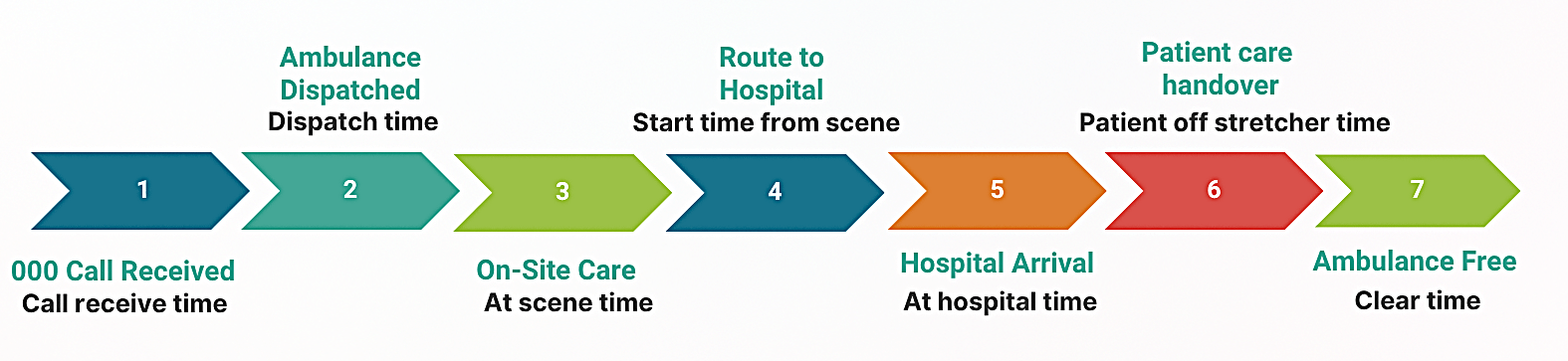}
\caption{Timestamped events in the EMS operational workflow. Seven key events recorded in the dataset (call received, ambulance dispatch, arrival at scene, begin transport, arrival at the medical facility, patient transfer, and back to service) are mapped to the corresponding operational intervals.}
\label{fig:EMSOperations}
\end{figure}
\textit{Total Service Time}: The full duration of ambulance involvement from the moment of initial dispatch until the ambulance is once again available, including travel to the scene, on-site operations, patient transport, and all interactions at the hospital. Fig. \ref{fig:EMSOperations} illustrates an overview of 000 call related ambulance activities. The time intervals shown quantify different stages of the EMS journey, including dispatch delay, which reflects response efficiency; patient transfer to hospital bed (off-stretcher) delay, which captures hospital ambulance ramping or patient offload delay; and the ambulance availability interval, which represents the total service turnaround time. These operational intervals form the basis for the subsequent analyses of ambulance ramping and EMS system performance in the paper.\\
\textit{Hospital Arrival Concurrency (Prior Arrivals 1h)}: A derived operational state feature computed for each case, counting the number of other ambulance arrivals at the same hospital in the 60 minutes preceding the case's own arrival. Higher values indicate cases arriving into more congested hospital environments. 
\subsection{Software and Analytical Tools }All data processing, statistical analysis, and visualisation for this study were performed in Python (version 3.11), with the analytical dataset stored and queried through a DuckDB (version 1.1) in-process analytical database. SQL queries were used for cohort construction, cross-tabulation, and aggregation across the dataset, while pandas (version 2.1) was used for downstream data manipulation. Visualisations were produced using matplotlib (version 3.8) and Seaborn (version 0.13). Custom Python functions handled the concurrency computation and the cross-correlation analysis. 
\subsection{Data Cleaning \& Assessments }
\begin{figure}[!t]
\centering
\includegraphics[width=0.85\columnwidth]{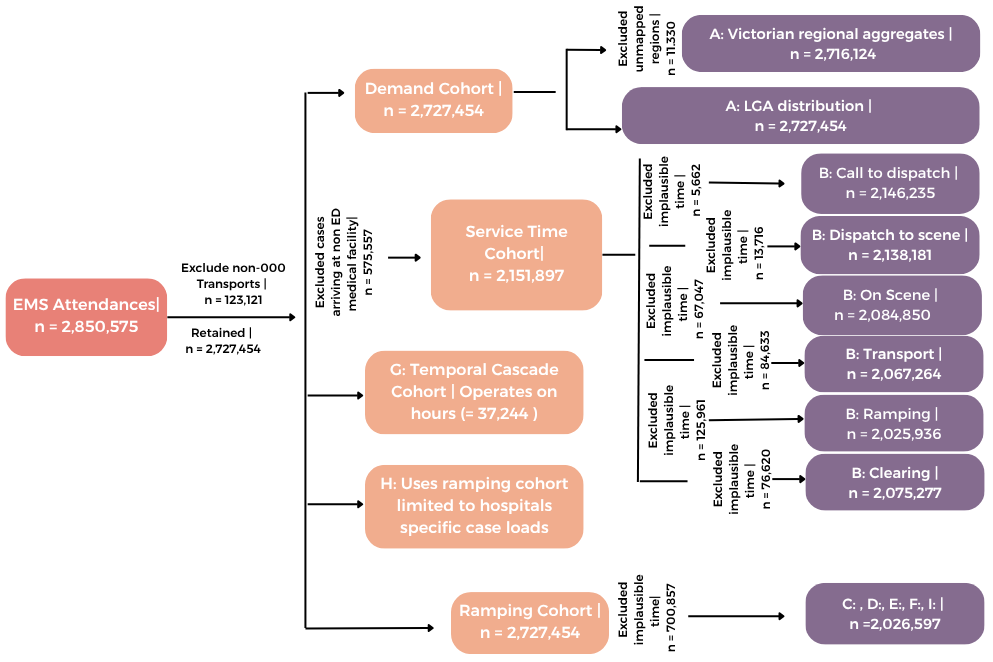}%{figures/flowchart.png}%{figures/Flowchart_after_dispatch_type_data.png}
%{figures/dataPreProcessing_colored_v1.2.PNG}
\caption{Derivation of the analytical cohorts, with the sample size for each results subsection. Cases not originating from public emergency calls were excluded throughout. Each cohort applies only the requirements of the analyses it supports.}
%{Flowchart illustrating cohort selection, exclusion criteria, and data pre-processing steps used to derive the final analytical cohort of 2.11 million records.}
\label{fig:datapreprocessing_flowchart}
\end{figure}

Each record in the dataset carries the pathway through which the case originated: emergency and non-emergency responses. The responses that do not originate from public emergency calls were excluded from all analyses as they do not represent community ambulance demand. This exclusion removed 123,121 cases (4.32\%). As the components of the ambulance service interval were recorded independently at different point in the total service time of a response, the availability and plausibility of these intervals may vary across measures within a case. Accordingly, analytic cohorts were defined separately for each analysis based on the data requirements of the relevant measure, rather than requiring complete and plausible data across all the components of the ambulance service interval. Fig. \ref{fig:datapreprocessing_flowchart} sets out the derivation and reports the cohort size for every results subsection. Demand analyses use all 2,727,454 retained dispatches, preserving the full operational load on the ambulance fleet. Ramping analyses are restricted to the 2,026,597 cases reaching an ED-capable facility with a plausible patient handover interval, since handover duration is measurable only at an emergency department interface; cases treated on scene, where transport was refused, or resulting in death at the scene are therefore excluded.
%Non-emergency planned transports are retained throughout, as they occupy the same handover capacity as unscheduled arrivals.
Ramping was calculated as the interval between ambulance arrival at the destination hospital and patient offload from the stretcher. Three categories of implausible value were excluded as timestamp errors: cases in which the offload timestamp preceded arrival, cases where the two timestamps were identical, and cases recording an interval under one minute, since an ambulance cannot arrive at an emergency department and complete offload within that time. A further 1,189 cases (0.06\%) recorded an interval exceeding 600 minutes and were likewise excluded. These are not extreme operational events: their median recorded duration was several days and the longest exceeded two months, durations incompatible with any plausible handover process. The resulting distribution of ramping duration is strongly right-skewed (Fisher Pearson skewness = 2.818, excess kurtosis = 12.488). The mean (44.15 minutes) exceeds the median (30.98 minutes) by 42.5\%, indicating that the arithmetic mean is disproportionately influenced by extreme values. The median was therefore adopted throughout as the measure of typical operational experience, and the 90th percentile (91.3 minutes, almost three times the median) as the tail metric characterising the worst-served cases. 

\section{Results and Discussion }
This section establishes where ambulance service time is consumed, how much capacity that loss represents, how it has changed across the study window, and where and under what conditions it concentrates. The analyses reported here are descriptive and correlational; associations should not be interpreted as causal. Importantly, the study window spans markedly different pandemic operating conditions. During the 2020 lockdown period, ambulance and ED demand was atypically suppressed, whereas the easing of restrictions from late 2021 was accompanied by recovering demand while pandemic-related infection-control procedures \cite{victorianhealth2023covidguidelines} remained in place. This combination may partly explain the pronounced AHL and ramping delays observed around 2021–22. Accordingly, these peaks should not necessarily be interpreted as evidence of underlying system deterioration, nor should the subsequent reductions in 2023–2024 be interpreted as evidence of system improvement or the effect of a specific intervention. In the absence of a pre-pandemic baseline, the longer-term direction of ramping performance cannot be established from these data alone.
\begin{figure}[!t]
\centering
\includegraphics[width=\columnwidth]{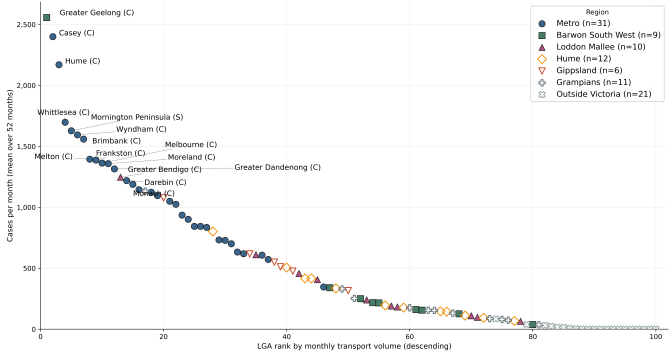}
%{figures/LGACaseLoadScatter.png}
\caption{LGA wise average monthly case loads over the full study period. first 15 LGAs with the most ambulance demand cases are labelled. Shape colors depict LGA regional mapping.}
\label{fig:caseLoadLGA}
\end{figure}
\begin{figure*}[ht!]
 \centering
 \includegraphics[width=\textwidth]{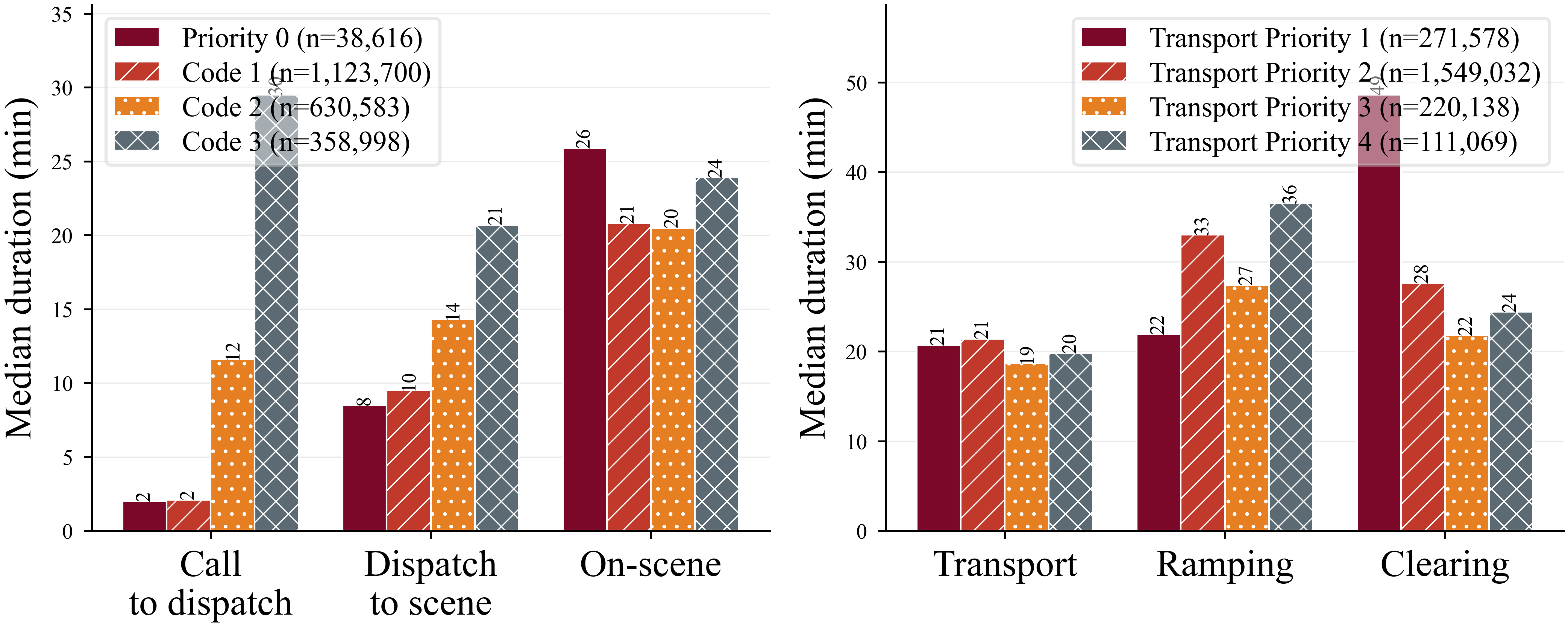}
 \caption{Duration of service time interval by patient acuity, shown as grouped bars. Intervals completing before the patient's on-scene assessment, stratified by dispatch code (left panel bars). The intervals occurring after it, stratified by transport priority. Components are not stacked, as percentiles are not additive; the two panels use different case groupings and are not to be read across.}
 \label{fig:composition_by_acuity}
 \end{figure*}
 \vspace{-1.75 ex}
\subsection{Ambulance Demand Distribution} Ambulance demand was characterised across Victorian LGAs over the study period. Mean monthly case load for the LGAs across the study window are ranked in descending order and labelled for the 15 highest demand volume areas. Demand is steeply concentrated: the highest volume LGA, Greater Geelong, averages 2,557 cases per month, while the 30th ranked LGA averages 728.9, where the volume declines to fewer than one case per month across lowest-ranked (based on case loads) LGAs, including unincorporated area of Victoria.
Notably, the highest volume LGA in the state is a regional centre rather than a metropolitan council. Greater Geelong (Barwon South West) exceeds the largest metropolitan LGA, Casey (2,400 average cases per month), indicating that high absolute demand is not exclusively a metropolitan phenomenon. Metropolitan councils nonetheless dominate the upper ranks, occupying 13 of the 15 highest demand volume positions, with Greater Geelong (Barwon South West), Greater Bendigo (Loddon Mallee) being the other regional centres appearing in that group. At the regional level, metropolitan Melbourne accounts for the majority of statewide demand, followed by Barwon South West, Loddon Mallee, Gippsland, Hume, and Grampians. This ordering reflects both the number of LGAs each region contains and the population each serves. The analysis reports absolute case counts and does not adjust for resident population. Note that the regional analyses exclude 11,330 cases (0.41\%) originating in LGAs outside the Victorian regional structure, principally interstate border councils (Fig. \ref{fig:caseLoadLGA}).
\vspace{-1.75 ex}
\begin{table}[!t]
\centering
\caption{Duration (minutes) of each service-time interval by patient acuity,
reported as median and 90th percentile (January 2020-March 2024). Intervals completing before the paramedic's transport assessment are stratified by dispatch code; those occurring after it by transport priority. Denominators differ because each interval is computed on the subset of cases with a plausible recorded value for that interval.} 
%Acuity groups are defined by the dispatch code assigned at call receipt.}
\label{tab:interval_decomposition}
\footnotesize
\setlength{\tabcolsep}{2.2pt}
\begin{tabular}{lllll}
\toprule
& \textbf{Priority 0} & \textbf{Code 1} & \textbf{Code 2} & \textbf{Code 3} \\
\textbf{Interval} & 38,616 & 1,123,700 & 630,583 & 358,998\\
& Med (p90) & Med (p90) & Med (p90) & Med (p90) \\
\midrule
Call to dispatch     &  2.0 (11.7) &  2.1 (15.2) & 11.6 (59.4) & 29.4 (94.5) \\
Dispatch to scene    &  8.5 (23.1) &  9.5 (21.1) & 14.3 (29.6) & 20.7 (55.9) \\
On-scene             & 25.9 (60.5) & 20.8 (38.0) & 20.5 (41.7) & 23.9 (45.8) \\
\midrule
%\textbf{Total}       & \textbf{134.2 (226.6)} & \textbf{125.8 (197.0)}
%                     & \textbf{130.3 (208.2)} & \textbf{147.2 (255.9)} \\
%\bottomrule
%\hline
\toprule
& \textbf{Transport 1} & \textbf{Transport 2} & \textbf{Transport 3} & \textbf{Transport 4} \\
\textbf{Interval} & 271,578 & 1,549,032 & 220,138 & 111,069 \\
& Med (p90) & Med (p90) & Med (p90) & Med (p90) \\
\midrule
ED Transport            & 20.7 (54.0) & 21.4 (47.0) & 18.7 (50.6) & 19.8 (44.8) \\
Ramping                 & 21.9 (59.9) & 33.0 (94.5) & 27.4 (83.2) & 36.5 (116.2) \\
Clearance               & 48.6 (91.6) & 27.6 (58.2) & 21.8 (48.3) & 24.4 (50.2) \\
\midrule
%\textbf{Total}       & \textbf{81.8 (206.2)} & \textbf{125.8 (197.0)}
%                     & \textbf{130.3 (208.2)} & \textbf{147.2 (255.9)} \\
%\bottomrule
\end{tabular}
\end{table}
\vspace{-1.75 ex}
\begin{figure}[!t]
\centering
\includegraphics[width=\columnwidth]{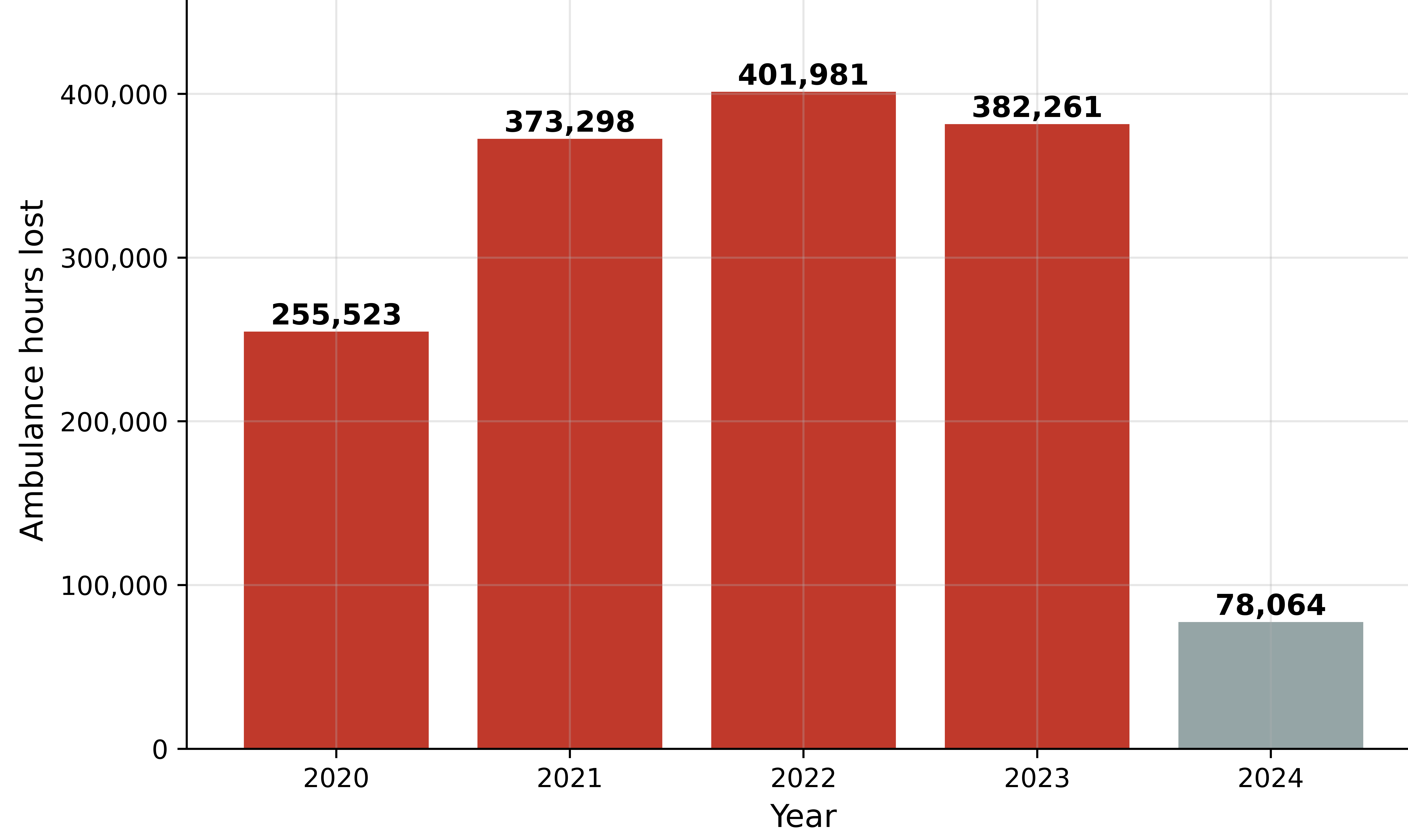}
\caption{Yearly trajectory of Ambulance Hours Lost (AHL) all the transports to ED between Jan 2020-March 2024. Total AHL hours per calendar year, annotated with the equivalent number of 10-hour ambulance shifts lost.}% Annual AHL rose from 265,894 hours in 2020 to a peak of 416,146 hours in 2022 (a 56.5\% increase) despite broadly stable annual case loads.
\label{fig:yearlyAHL}
\end{figure}
\subsection{Decomposition of Service Time by Dispatch Code and Transport Priority} The total ambulance service interval comprises multiple sequential operational phases, whose durations may reflect both patient-related and system-related factors. Decomposing this interval and examining its association with patient acuity provides a basis for identifying the components that may be most amenable to operational improvement. Each interval is stratified by the assessment that governed it. The call to dispatch and dispatch to scene are completed before the attending paramedic(s) assigns a transport priority to the patient during on-scene assessment, and are therefore stratified by the dispatch code assigned at call receipt. The transport, ramping and clearing intervals occur after that assessment which are stratified by the paramedic assigned transport priority (Fig. \ref{fig:composition_by_acuity}). 
Call to dispatch varies sharply by a wide margin, with a median of 2.0 minutes for Priority 0 \footnote{Priority 0 is intended as a diagnostic subgroup of Code 1 and remains included in the overall Code 1 category.} cases against 29.5 minutes for Code 3, a 14.75 fold difference. Dispatch to scene follows the same direction less steeply (8.5 against 20.7 minutes, a factor of 2.44). Clearing time varies inversely with transport urgency, from 48.6 minutes for a transport priority 1 to 21.8 and 24.4 minutes for transport priority 3 and 4 patients respectively, reflecting more preparation time required following a critical patient handover. Transport time is effectively invariant, ranging only from median of 18.7 to 21.4 minutes across all four transport priority groups. The ramping delay (handover delay) ranges from median ramping of 21.9 minutes for transport priority 1 but 33.0, 27.4 and 36.5 minutes for priorities 2, 3 and 4 respectively, a factor of 1.66 overall, with no ordering whatsoever among the three lower priorities. The only acuity signal present in the ramping delay is the expediting of priority 1 arrivals; beyond this, how long a patient waits at the hospital door bears no relation to their assessed urgency. This is consistent with a triage protection mechanism whereby the highest acuity patients appear relatively insulated from ambulance turnaround delays, supporting the finding that preferential protection is concentrated at the most critical acuity level \cite{li2026empirical}. Priority 4 transports show the longest delays of all groups (90th percentile of 116.2 minutes, compared with 59.9 minutes for priority 1), indicating that these cases are the most severely impacted (Table~\ref{tab:interval_decomposition}).
Beyond a modest expediting of the most critical acuity group, ramping delay shows relatively limited and non-monotonic variation with patient acuity. The distinction is important for intervention design: delays determined by patient acuity would be less amenable to operational intervention whereas delays that are largely independent of acuity may reflect modifiable operational factors. Whether these delays vary across hospitals and across hour of the day, is investigated further in the following sections.
%%CHECKING THIS ========================
\subsection{Ambulance Hours Lost: System and Regional Burden}\label{subsec:AHL_SystemBurden} Over the 4.25 year study window, total cumulative AHL was 1,491,127 hours-equating to 149,113 lost 10-hour shifts, or roughly 96 ambulance-shift equivalents per day. This represents roughly 96 ambulances being held at hospitals for a full 10-hour shift, every day throughout the study period. Regional attribution of AHL (Table \ref{tab:Per_Region_AHL}) shows that Metropolitan Melbourne accounted for 76.7\% of total AHL (73.6 shifts lost per day), while the five regional zones collectively accounted for 23.3\% of total AHL, distributed as Barwon South West (6.1\%), Loddon Mallee and Hume (4.8\%), Gippsland (4.1\%), and Grampians (3.5\%). Mean per case ramping was highest in the Metro (46.8 min) and Hume (41.3 min) regions, while the four remaining regional Victoria zones showed tightly clustered per case ramping (35-37 min). The Metropolitan concentration of AHL modestly exceeds the region's share of total ED cases (72.3\% AHL cohort share), indicating both a volume driven concentration and a consistent mean per case ramping in the Metro region.
\begin{table}[!t]
\centering
\caption{Regional ambulance hours lost (AHL) by hospital region across the study window.}
\label{tab:Per_Region_AHL}
\footnotesize
\begin{tabular}{lrrrr}
\toprule
\textbf{Region} &
\textbf{Cases} &
\textbf{AHL (h)} &
\textbf{Mean} &
\textbf{Shift}\\
&
&
&
\textbf{Ramp (min)} &
\textbf{Lost/day}\\
\midrule
Metro                & 1,465,470 & 1,143,320.9 & 46.8 & 73.6 \\
Barwon SW            &   149,332 &    91,513.9 & 36.8 & 5.9 \\
Hume                 &   105,094 &    72,315.4 & 41.3 & 4.7 \\
Loddon Mallee        &   121,119 &    71,660.9 & 35.5 & 4.6 \\
Gippsland            &   102,497 &    60,640.7 & 35.5 & 3.9 \\
Grampians            &    83,085 &    51,674.8 & 37.3 & 3.3 \\
\bottomrule
\end{tabular}
\end{table}

\begin{table}[!t]
\centering
\caption{Yearly trajectory of Ambulance Hours Lost (AHL) across the study window. The cohort comprises all transports taken place during 4.25 years to the 59 hospitals with ED. Shifts lost are expressed in 10-hour shift equivalents. The 2024 row reflects partial year data (January–March, 91 days) and is reported with its daily average rate for comparability with full calendar years.}
\label{tab:Yearly_AHL}
\footnotesize
\begin{tabular}{lrrrr}
\toprule
\textbf{Year} &
\textbf{n\_cases} &
\textbf{AHL} &
\textbf{Mean} &
\textbf{Avg Shifts}\\
&
&
\textbf{Ramp(h)} &
\textbf{Ramp/day}&
\textbf{Lost/day}\\
\midrule
2020    & 479,203   & 255,522.6  & 32.0  & 69.8   \\
2021    & 509,647    & 373,298.1   & 43.9  & 102.3   \\
2022    & 461,418    & 401,980.7  & 52.3  & 110.1  \\
2023    & 471,688    & 382,261.1   & 48.6  & 104.7   \\
2024    & 104,641     & 78,064.1    & 44.8  & 84.9   \\
\bottomrule
\end{tabular}
\end{table}
\subsection{Yearly Trajectory of Ambulance Hours Lost }\label{subsec:yearly_AHL}
Decomposing the full-period AHL total by calendar year reveals a substantial and sustained escalation across the study window (Fig. \ref{fig:yearlyAHL}). Annual AHL rose from 255,523 hours in 2020 (69.8 lost 10-hours shift per day) to 373,298 hours in 2021 (102.3 lost shifts/day) and peaked at 401,981 hours in 2022 (110.1 shifts/day), a 57.3\% increase over two years, before easing marginally to 382,261 hours in 2023 (104.7 shifts/day). The partial 2024 window (January-March) recorded 78,064 hours (84.9 shifts/day) is consistent with the elevated post-2020 operating level rather than attaining the 2020 level (Table \ref{tab:Yearly_AHL}). 

Critically, this escalation is not explained by demand growth. Annual ED transports were broadly stable; 479,203 cases in 2020, 509,647 in 2021, 461,418 in 2022, and 471,688 in 2023, while mean per case ramping rose from 32.0 minutes (2020) to 52.3 minutes (2022), a 63\% increase. Andrew et al.\cite{1c0d44cf377646259d12e84bff7b56ac} provided Victoria-specific evidence of this phenomenon: despite a reduction in ambulance call volumes during the 2020 lockdown, response and case times worsened, with the deterioration driven predominantly by extended hospital transfer times. In another retrospective cohort study of the 25 largest Queensland public hospitals, rising ramping is explicitly attributed to access block rather than demand \cite{yoon2026association}. The AHL trajectory is not proportionally driven by the case load size, indicating changing conditions at the ambulance ED interface rather than growth in transported demand.
%or per case handover severity rather than demand growth.
\begin{figure*}[!t]
\centering
\includegraphics[width=\textwidth]{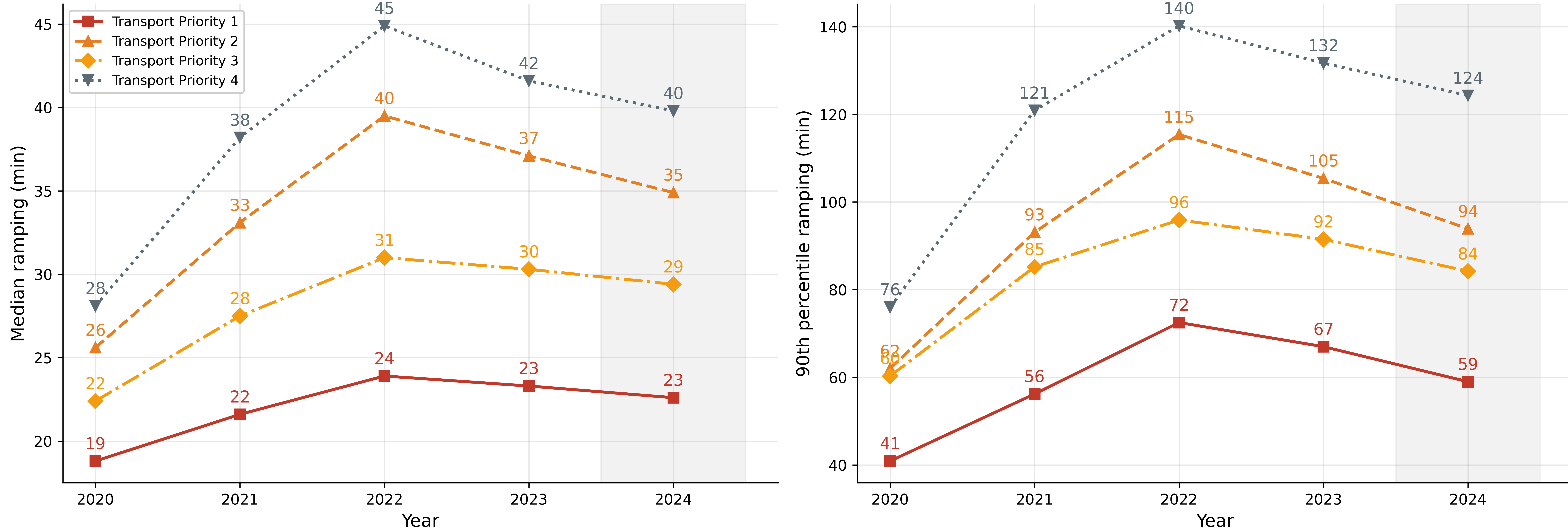}%{figures/yearlyRampingTrajectory_highest_transport_priority_priority2_3_4_reconciled.PNG}
\caption{Yearly ramping trajectory by patient's transport priority code: median (left) and 90th percentile (right). Ramping times increased across all transport priorities from 2020 to a peak in 2022, with partial easing in 2023; Transport Priority 4 carries the heaviest tail throughout. 2024 (shaded) covers January–March only}
\label{fig:yearly_ramping_arr_code}
\end{figure*}
\begin{figure}[!ht]
\centering
\includegraphics[width=\columnwidth]{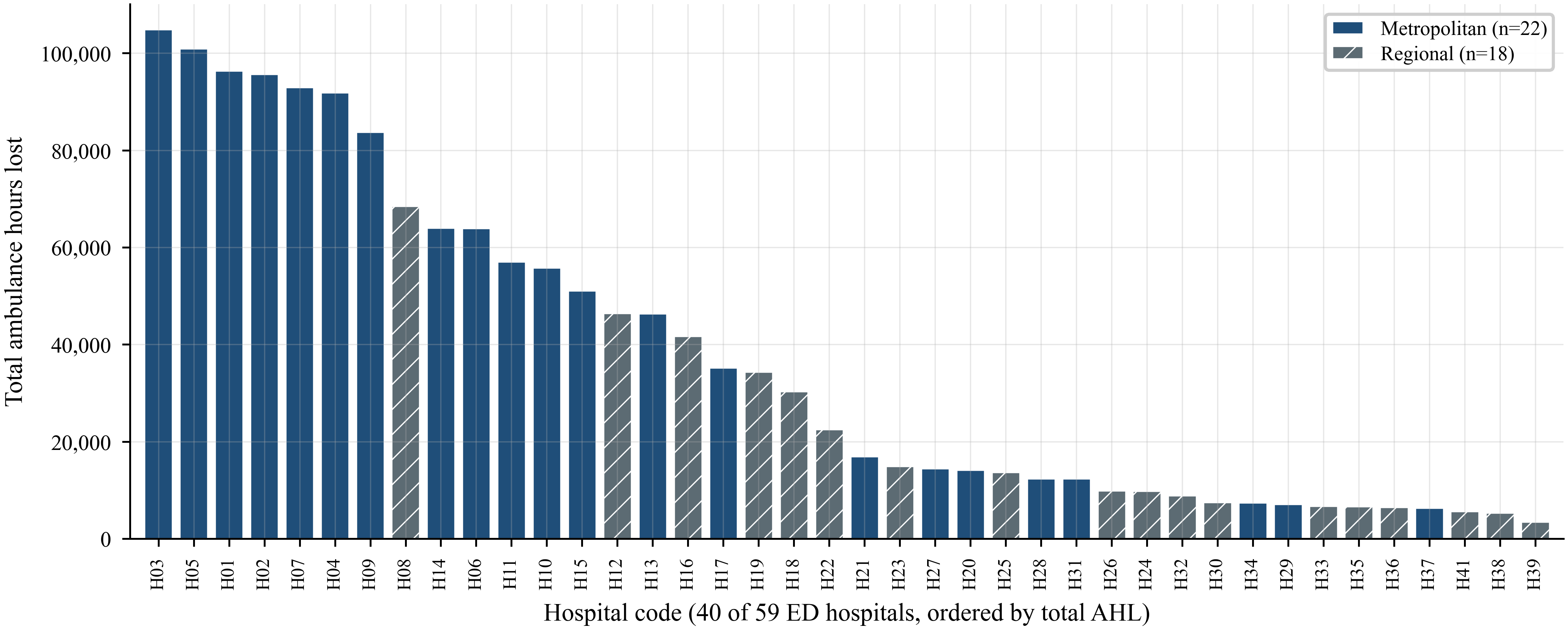}
\caption{Hospital wise ambulance hours lost across the study period Jan 2020 - Mar 2024, demonstrates that the first 15 medical facilities bear the 75\% AHL while their case share is 68.6\%. Out of those 15 facilities, 13 are in the Metro (Refer Fig. \ref{fig:ParetoAHL}) region, Victoria.}
\label{fig:HospitalWiseAHL}
\end{figure}
\subsection{Yearly Composition and Ramping Trajectory by Transport Priority } \label{subsec:yearly_compo_AHL_patientAcuity}The yearly distribution of transported cases by transport priorities (Table \ref{tab:yearly_transport_arrival_priority_volume}) reveals a sustained shift in transport priorities composition across the study window, alongside broadly stable total volume. Annual transports ranged between 461,418 (2022) and 509,647 (2021) across the four full calendar years 2020-23. Within this stable envelope, transport priority 1 cases rose from 47,957 in 2020 to 62,745 in 2023, increasing from 10.0\% to 13.3\% of annual volume, while lower transport priority category declined over the same period, transport priority 2 from 353,460 to 341,457, transport priority 3 from 49,880 to 46,178, and priority 4 from 27,906 to 21,308. 
\begin{table}[!t]
\centering
\caption{Yearly distribution of cases by transport priority(1-4) from Jan 2020-March 2024.}
\label{tab:yearly_transport_arrival_priority_volume}
\renewcommand{\arraystretch}{1.15}
\begin{tabular}{lrrrrr}
\toprule
\textbf{Year} & \textbf{Priority 1} & \textbf{Priority 2} & \textbf{Priority 3} & \textbf{Priority 4} & \textbf{Total} \\
\midrule
2020 & 47,957 & 353,460 & 49,880 & 27,906 & 479,203 \\
2021 & 53,378 & 371,670 & 54,366 & 30,233 & 509,647\\
2022 & 57,565 & 338,261 & 45,904 & 19,688 & 461,418\\
2023 & 62,745 & 341,457 & 46,178 & 21,308 & 471,688\\
2024 & 14,026 & 73,365  & 11,490 & 5,760 & 104,641\\
%\midrule
%\textbf{Total} & \textbf{271,966} & \textbf{1,551,009} & \textbf{270,670} & \textbf{234,934} & \textbf{186,796} \\
\bottomrule
\end{tabular}
\end{table}
The yearly ramping trajectory, stratified by transport priority (Fig. \ref{fig:yearly_ramping_arr_code}), follows the system level pattern established in Section \ref{subsec:yearly_AHL}: deterioration from 2020 to a 2022 peak, followed by partial easing in 2023 that remains well above the 2022. For transport priority 1 cases, 90th percentile ramping rose from 41 minutes (2020) to 72 minutes (2022), an 75.6\% increase, and remains elevated at 67 minutes in 2023. The same shape holds across transport priorities 2, 3, and 4, where priority 4 consistently carrying the heaviest tail (p90 of 76, 121, 140, and 132 minutes across 2020–23). The acuity gradient observed in the whole period analysis is therefore stable across years: at every point in the window, lower acuity arrivals wait longer at handover than time critical arrivals, and the gap widened in absolute terms as the system deteriorated. Taken together, the compositional shift and ramping trajectory compounded one another: as an increasing share of arrivals fell under the highest transport priority, the category hospitals expedite, the delay experienced by the remaining majority of the queueable caseload, comprising lower transport priorities (priorities 3 and 4), lengthened substantially through 2022.
% \label{fig:yearly_dispatch_arr_code}
% \end{figure*}
\subsection{Hospital Level Concentration of AHL } Following the system-level analysis of AHL, a hospital-level analysis was conducted to provide a more granular assessment of its distribution and concentration across individual hospitals. The 59 hospitals with ED in the analytical cohort have shown the highest concentration of AHL. The cumulative AHL total was 1,491,127 hours distributed across 2,026,597 cases over the study window. 
Fig. \ref{fig:HospitalWiseAHL} presents the 40 hospitals ranked by their total AHL contributions, with the bar colors indicating whether each hospital is metropolitan or regional. The distribution is heavily skewed. The largest contributor accounts for approximately 109,530 hours (7.03\% of the total AHL), the tenth largest for 57.68\%, and the forty largest for 98.7\%. Of the 15 highest contributing hospitals, 13 are in the Metropolitan Melbourne region; the exceptions are H08 (rank 8, Regional) and H12 (rank 14, Regional). All six regions are represented within the 20 highest AHL contributing hospital.
\begin{figure}[t!]
\centering
\includegraphics[width=\columnwidth]{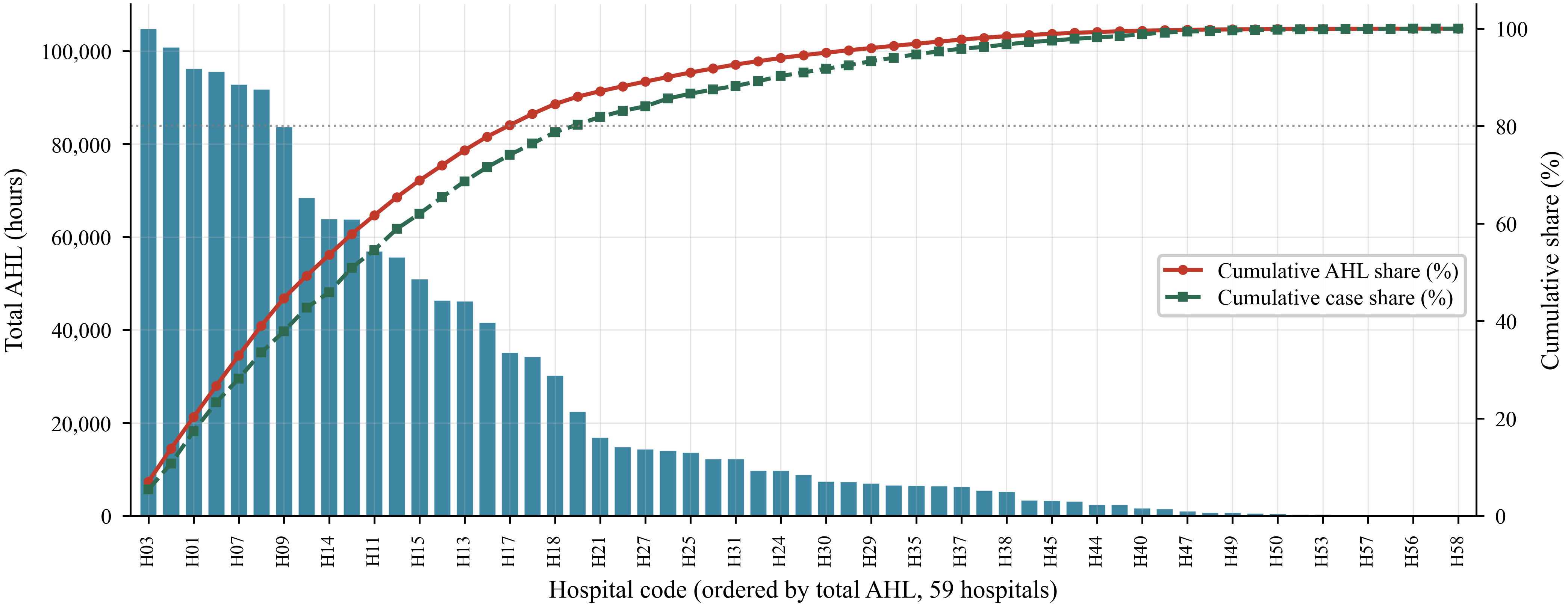} %{figures/Pareto_AHL.pdf}
\caption{Pareto analysis: The medical facilities ranked 41-59 contribute a combined ~2-3\% of the total AHL, their bars are near-invisible. The facilities showing cumulative AHL share (solid line) and cumulative case share (dashed line) as functions of rank.}
\label{fig:ParetoAHL}
\end{figure}
Concentration alone, however, partly reflects case volume: hospitals receiving more ambulances will accumulate more handover time. The Pareto analysis in Fig. \ref{fig:ParetoAHL} therefore compares cumulative AHL share against cumulative case share across the ranked medical facilities. The two curves separate consistently across the upper ranks. The gap between the two curves, which shows the disproportionality of AHL share over case share, is visible across the top 45 ranks and closes over after rank 45. Table \ref{tab:cumulative_ahl_share} summarises the cumulative share statistics at key ranks. 
\begin{table}[ht]
\centering
\caption{Cumulative hospital contributions to AHL across 59 EDs, assessed at key ranking positions. Hospitals are ordered by decreasing total AHL contribution. Difference (percentage points, pp) quantifies the excess of cumulative AHL share over cumulative case share.}
\label{tab:cumulative_ahl_share}
\begin{tabular}{rrrr}
\toprule
\textbf{Rank} &
\textbf{Cum. AHL (\%)} &
\textbf{Cum. Cases (\%)} &
\textbf{Difference (pp)} \\
\midrule
5  & 32.9 & 28.2 & +4.7 \\
10 & 57.8 & 50.9 & +6.9 \\
15 & 75.0 & 68.6 & +6.4 \\
20 & 86.0 & 80.3 & +5.7 \\
25 & 91.0 & 86.6 & +4.3 \\
30 & 94.5 & 91.0 & +3.5 \\
40 & 98.7 & 97.2 & +1.5 \\
50 & 99.9 & 99.7 & +0.2 \\
\bottomrule
\end{tabular}
%\vspace{0.5em}
\end{table}
The 10 hospitals with highest AHL accounted for 57.8\% of total AHL while receiving 50.9\% of transport cases which is a 6.9 percentage point excess. The top 15 hospitals contributed 75.0\% of AHL from 68.6\% of cases (a 6.4 percentage point excess), and the top 20 hospitals contributed 86.0\% of AHL from 80.3\% of cases (a 5.7 percentage point excess). The excess is greatest at rank 10, declines steadily thereafter to 5.7 percentage points at rank 20 and 4.3 at rank 25, and is negligible by rank 50, where the cumulative shares converge (99.9\% and 99.7\% respectively). Disproportionate AHL contribution is therefore real but bounded to approximately the first 25 to 30 hospitals, rather than being distributed across the full medical facilities.
This pattern is also evident at the level of individual facilities. Among the ten highest contributors, H14 generates 4.29\% (63,917.8 hours) of statewide AHL from 3.15\% of transports (63,910), a ratio of 1.36, whereas H06 generates 4.28\% (63,863.4 hours) from 5.04\% of transports (102,117), a ratio of 0.85. The two hospitals appear next to each other in total AHL, but they arrive there via different paths: one due to higher handover time per case, the other driven solely by case volume. Across the ten highest contributors this ratio ranges from 0.85 to 1.36, indicating that similar aggregate burdens can arise from substantially different operational profiles.

\begin{table}[!t]
\centering
\caption{Cross-correlation between hourly demand and arrivals with hourly mean ramping.}
\label{tab:cross_correlation_lags}
\small
\setlength{\tabcolsep}{4pt}
\begin{tabular}{c r r @{\qquad} c r r}
\toprule
\textbf{Lag} & \textbf{Demand} & \textbf{Arrivals} &
\textbf{Lag} & \textbf{Demand} & \textbf{Arrivals} \\
\midrule
0 &  0.2372 &  0.3168           & 7  &  0.2031 &  0.1165 \\
1 &  0.3130 &  0.3525           & 8  &  0.1274 &  0.0349 \\
2 &  0.3526 & \textbf{0.3639}   & 9  &  0.0452 & -0.0437 \\
3 &  \textbf{0.3652} &  0.3524  & 10 &  -0.0335 & -0.1289 \\
4 &  0.3532 &  0.3181           & 11 & -0.1171 & -0.2086 \\
5 &  0.3209 &  0.2643           & 12 & -0.1974 & -0.2674 \\
6 &  0.2688 &  0.1942           & -- & -- & -- \\
\bottomrule
\end{tabular}
\end{table}
%\vspace{-2.0ex}

\subsection{Temporal Cascade of Demand, Arrivals, and Ramping }
\begin{figure}[!t]
\centering
\includegraphics[width=\columnwidth]{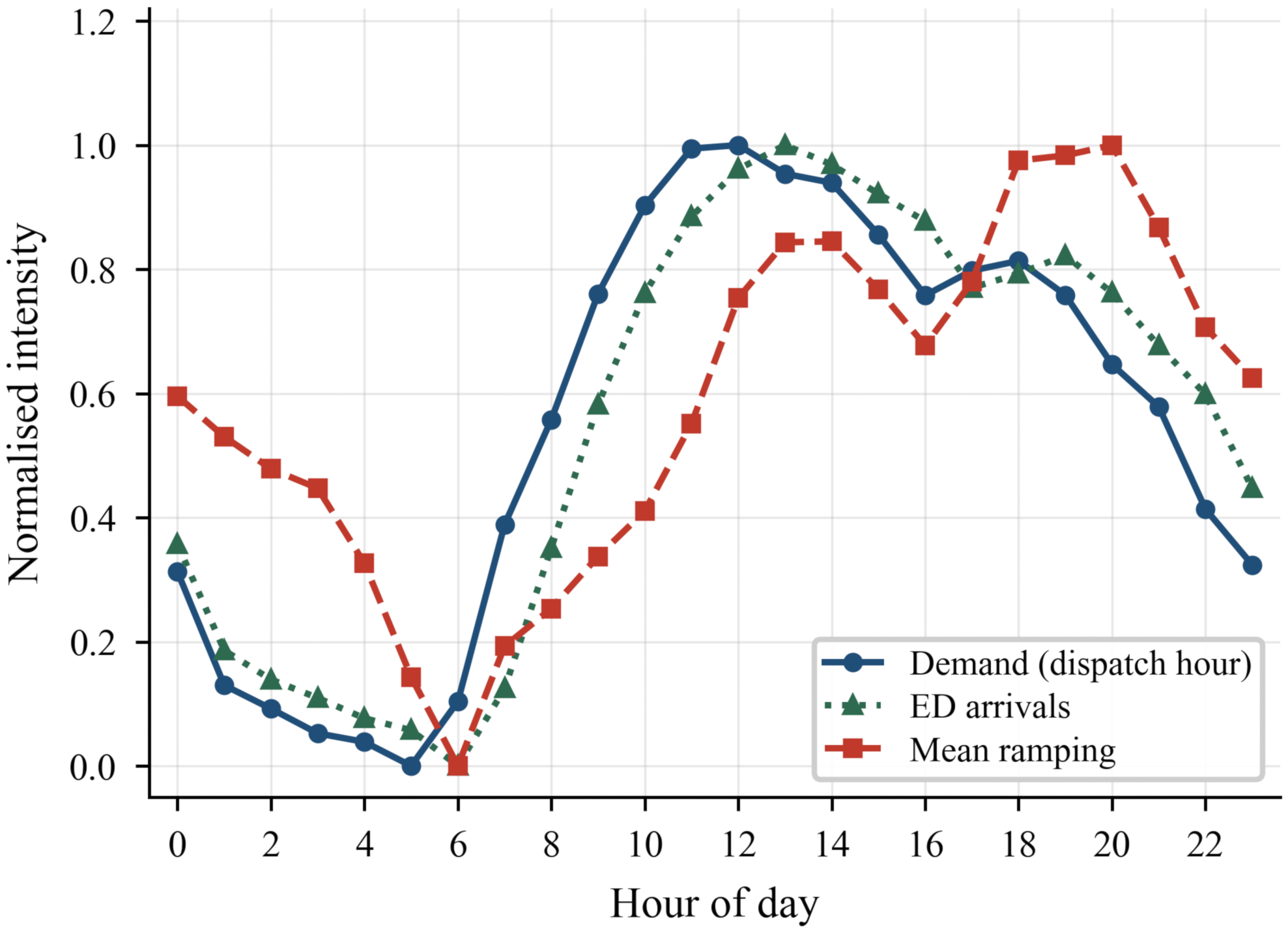}
\caption{Hour of day profile of ambulance demand, ED arrivals and mean ramping duration, averaged across the study window. Demand peak 130,620 cases per hour of day, arrivals peak 125,583, mean ramping peaks 50.65 minutes.}
\label{fig:HODCascaseOverlapDispatchRamping}
\end{figure}

To investigate how ambulance demand and ED arrivals propagates into ramping, three hourly series were constructed: case loads indexed by ambulance dispatch hour, emergency department arrivals indexed by hospital arrival hour, and mean ramping duration also indexed by arrival hour. Demand and arrivals comprise the same cohort counted at two points in the ambulance journey; ramping is computed on the subset of ED arrivals with a recorded ramping/handover delay. Fig. \ref{fig:HODCascaseOverlapDispatchRamping} presents the three series averaged across the study window, each min-max normalised within itself for visual comparison. Demands rise from a minimum of 35,442 cases per hour of day at 05:00 to a peak of 130,620 at 12:00. ED arrivals follow the same shape displaced by an hour, from 36,124 at 06:00 to 125,583 at 13:00. Mean ramping follows a different trajectory: it reaches its minimum of 30.30 minutes at 06:00, coinciding with the arrival trough, then rises through the day to a plateau of 50.26 to 50.65 minutes spanning 18:00 to 20:00. The three troughs align closely, whereas the ramping peak lags the arrival peak by seven hours and the demand peak by eight. However, these differences in peak timing describe the typical daily pattern and should not be interpreted directly as a complete causal delay.
%\vspace{-1.0 ex}
Table \ref{tab:cross_correlation_lags} presents cross-correlation functions for both predictors against ramping across the 37,244 hours for which all three series were observed. In the raw hourly data, demand attains its maximum association with ramping at a lag of three hours ($r = 0.365$). Arrivals attain their maximum one hour earlier, at a lag of 2 hours ($r = 0.364$). 
%The correlations remained relatively similar across lags of two to four hours, indicating a broad rather than sharply defined operational timescale.
Across lags one to four, correlations range from 0.313 to 0.365 for demand and from 0.318 to 0.364 for arrivals, a span of 0.052 and 0.046 respectively. The association is therefore broad rather than concentrated at a single lag and effectively flat across this window rather than identifying a single sharply defined lag. The one hour separation between the two peak lags is visible in the daily curves; this displacement is consistent with the expected progression from dispatch to hospital arrival. The two predictors attain near identical peak correlations with ramping (0.365 for demand at three hours, 0.364 for arrivals at two), indicating that arrivals at the hospital door convey no additional information about handover duration beyond that already present in dispatch volume an hour earlier.
%Because case load/dispatch is available at the time of resource allocation, whereas hospital arrivals are observed only upon ambulance arrival at the hospital, thus former appears more promising in terms of providing earlier actionable signal to analyse ramping patterns. 
These results provide an apparent moderate association between higher ambulance activity and subsequent ramping at the raw hourly level, but they also indicate that ambulance activity alone explains only a limited proportion of the variation in ramping. Importantly, the strong association observed in the averaged hourly profiles should be interpreted cautiously. The peak correlations calculated from the 24-point averaged profiles were substantially higher, reaching $r=0.899$ for demand at a three-hour lag and $r=0.899$ for arrivals at a two-hour lag. 
%However, both demand and arrivals exhibit strong diurnal structure, with hour of day explaining 80.6\% and 81.4\% of their respective variances, compared with only 20.1\% for ramping. 
The high correlations between the averaged profiles therefore largely capture the fact that ambulance activity and ramping vary systematically over the course of the day, with ramping occurring several hours after the corresponding increase in demand and ED arrivals. Importantly, the hourly averaged profiles smooth substantial variability associated with individual ED circumstances. Thus, the strong association observed in the averaged profiles should be interpreted as evidence of a clear daily temporal relationship rather than as evidence that ambulance demand alone determines ramping at the individual hour level. In particular, the averaging process reduces short term fluctuations arising from individual ED conditions and therefore highlights the component of ramping associated with overall ambulance demand.
%rather than demonstrating a strong relationship between individual hourly deviations in ambulance activity and ramping. 
%After removing the hour of day effect, the association with ramping decreased substantially. The peak correlation was only $r=0.075$ for demand and $r=0.076$ for arrivals, occurring at lags of seven and six hours, respectively. Further adjustment for hour of week and the long-term trend (a least-square linear time trend) produced similarly weak associations, with peak correlations of $r=0.129$ for demand and $r=0.127$ for arrivals, both at a lag of seven and six hours, respectively. The persistence of a strong demand-arrival relationship after temporal adjustment provides an important control: demand and arrivals retained a peak correlation of $r=0.795$ after removing the hour-of-day effect, confirming that the adjustment did not simply eliminate temporal dependence from all series.
%(Refer Table \ref{tab:cross_correlation_schemes}. 
These results demonstrate a clear daily temporal progression in which ambulance demand is followed by ED arrivals and subsequently by ramping. Ambulance demand and ED arrivals exhibit a clear and consistent daily pattern, with arrivals occurring approximately one hour after dispatch demand, while ramping reaches its daily maximum several hours later. 
%At the individual-hour level, however, the association between ambulance activity and subsequent ramping is considerably weaker, and it becomes minimal after accounting for regular temporal patterns. 
This suggests that the apparent daily cascade is driven predominantly by shared time-of-day and operational structure, while short term variation in ramping is influenced by additional hospital side factors, such as capacity constraints, staffing, bed availability, patient flow, and other sources of ED workload not captured by ambulance arrivals alone.
\begin{figure*}[ht]
\centering
\includegraphics[width=\textwidth]{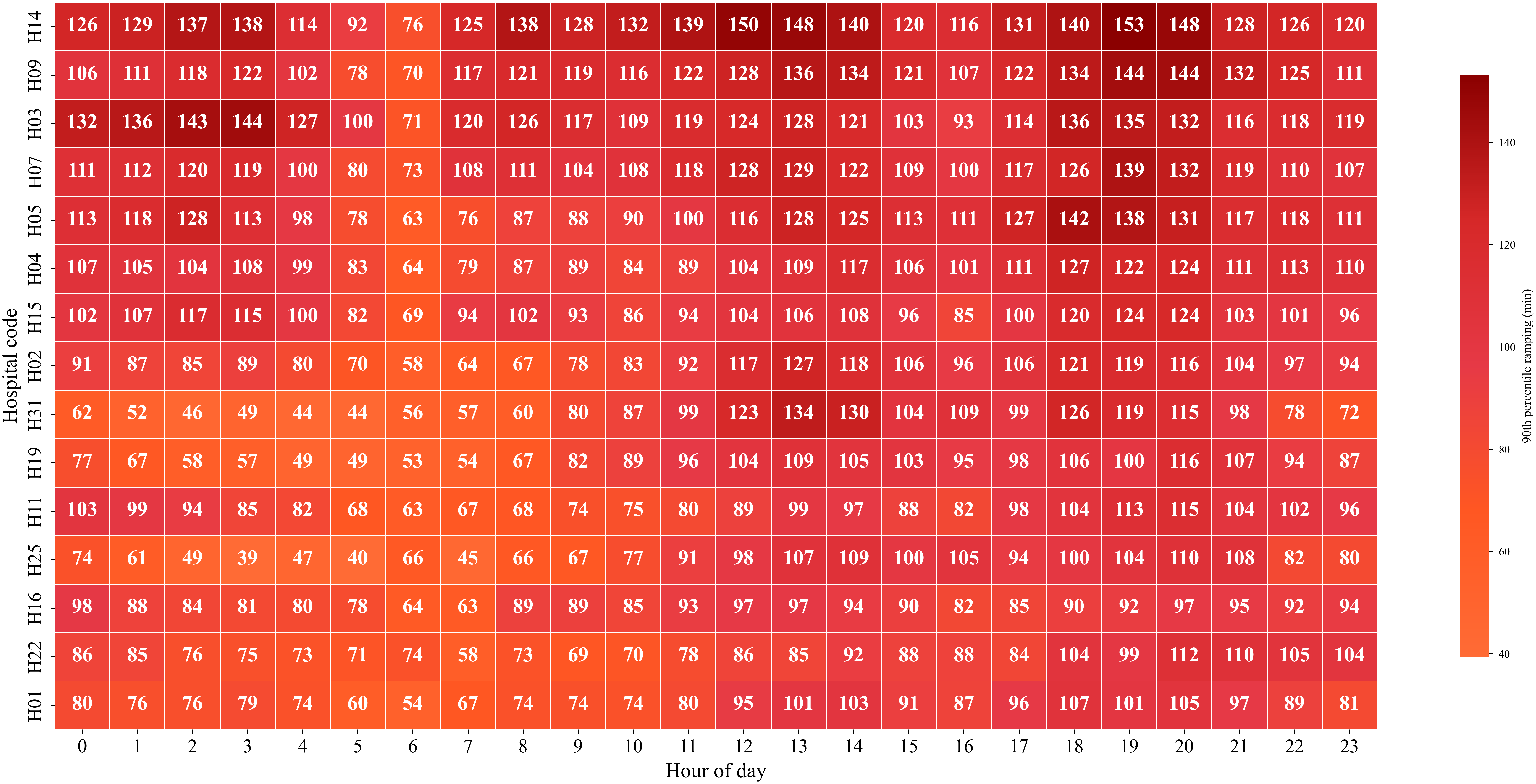}
\caption{Hourly 90th percentile (p90) ambulance ramping time (minutes) for the 15 hospitals with the highest overall p90 ramping. Hospitals are anonymised and ranked in descending order of their overall p90 ramping across the study period. Cell values and colour intensity represent the hourly p90 ramping duration, facilitating comparison of temporal congestion patterns among the most severely affected hospitals.}
\label{fig:hospP90RampingOverall}
\end{figure*}
\subsection{Hospital Level Temporal Heterogeneity }
\begin{figure}[ht!]
%\left
\includegraphics[width=\columnwidth]{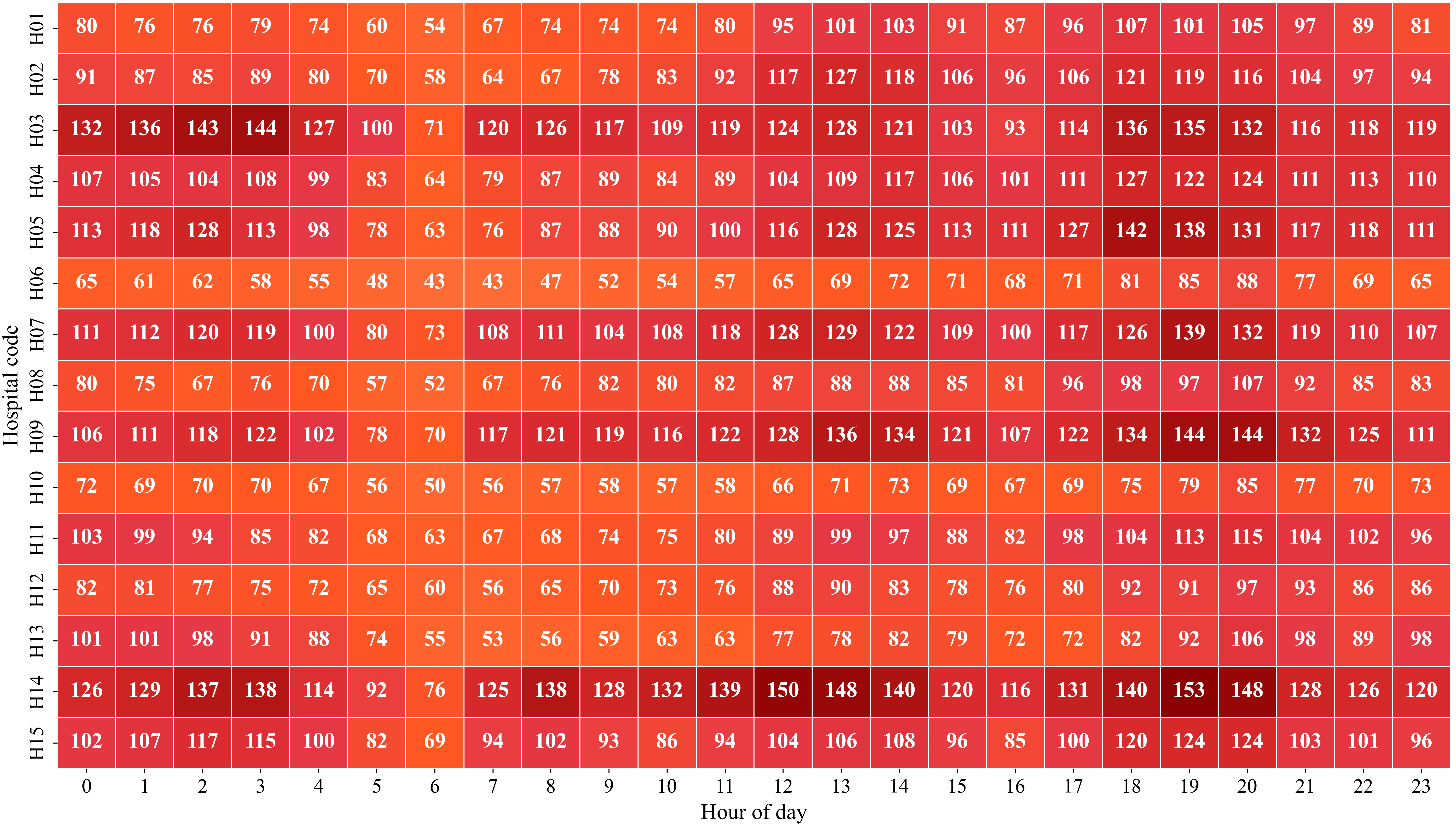}
\caption{Hourly 90th percentile (p90) ambulance ramping time (minutes) for the 15 hospitals with the highest ambulance case loads. Hospitals (H01-H15) are anonymised and ordered by decreasing ambulance case loads. Cell values and colour intensity represent the hourly p90 ramping duration, highlighting temporal variation in severe ramping across high demand hospitals.}
\label{fig:hospP90Ramping}
\end{figure}

Further analyses were conducted to determine whether hospitals in the ED cohort exhibit systematic differences in the onset, timing, and intensity of ramping that could inform hospital-specific forecasting or adaptive ambulance routing.
We performed two complementary analyses to characterise temporal heterogeneity at the hospital level (investigating hospital by hour of day) among the first 15 hospitals ranked by case load. Fig. \ref{fig:hospP90RampingOverall}, \ref{fig:hospP90Ramping} display heatmaps of the 90th percentile of ramping duration across these 15 hospitals (with rows ordered by ramping duration in Fig. \ref{fig:hospP90RampingOverall} and by case loads in Fig. \ref{fig:hospP90Ramping}) and the 24 hours of the day (columns).
Fig. \ref{fig:hospP90RampingOverall} where hospitals are arranged by their 90th percentile ramping duration, illustrates that hospitals do not experience identical temporal ramping profiles. Although nearly all hospitals exhibit a similar diurnal pattern, the magnitude and timing of peak congestion differ substantially.
Several observations emerge.
\begin{figure*}[!t]
\centering
\includegraphics[width=\textwidth]{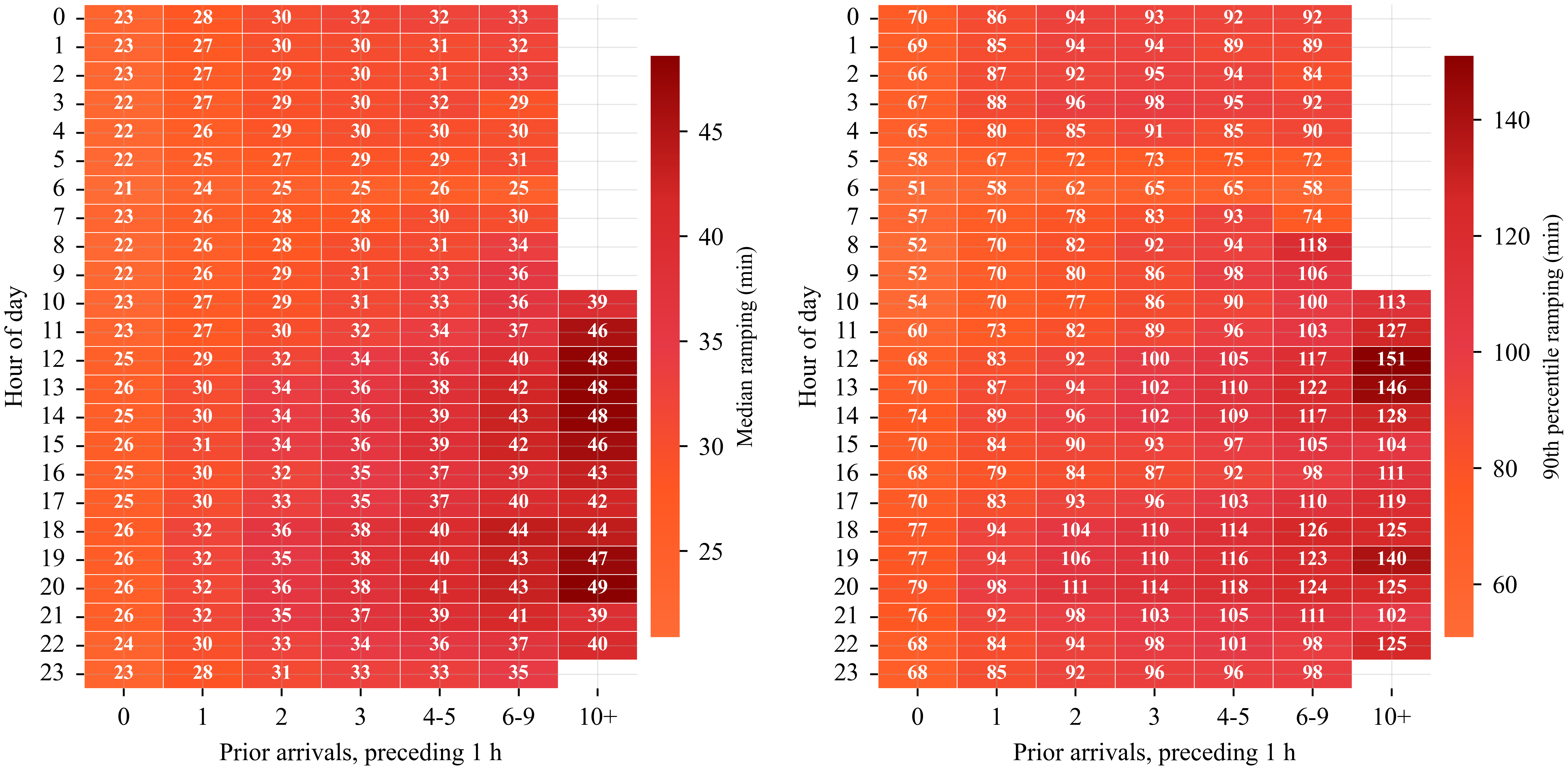}%{figures/concurrency_heat.pdf}
\caption{Median and the 90th percentile ramping duration stratified by hour of day (rows) and concurrency bin (columns), across the ED cohort. Concurrency is defined as the number of other ambulance arrivals at the same hospital in the 60 minutes preceding each case's own arrival. Colour intensity reflects their ramping magnitude. The joint gradient, rising ramping from left to right within any row AND from top to bottom within any column, indicates that hospital arrival concurrency and hour of day each contribute independently to ramping outcomes; neither is an artifact of the other. Cells with fewer than 50 cases are masked in white.} %(rare combinations, primarily off peak hours with high concurrency)
\label{fig:OneHourBeforeArrivalRampingHeatMap}
\end{figure*}
First, most hospitals show their lowest p90 ramping during the early morning hours (approximately 05:00–06:00), with values commonly falling between 44 and 104 minutes. This suggests that emergency departments begin each day with relatively lower congestion following overnight reductions in patient occupancy.
Second, ramping increases progressively through the late morning, reaching elevated levels during the afternoon and evening. For most hospitals, the highest p90 ramping occurs between 12:00 and 20:00, indicating a common system wide period of congestion.
Third, despite sharing similar daily trends, the magnitude of congestion varies markedly between hospitals. For example, hospitals such as H03, H09 and H14, frequently record p90 ramping exceeding 130 minutes, whereas hospitals such as H01, H11, H16, H19, H22, H25 remain consistently below or slightly more than 100 minutes throughout the day despite also receiving substantial ambulance case loads (H01 being the hospital with the most case loads). Hospital such as H31 display moderate overnight delays followed by sharp increases during midday and afternoon periods. Such inclusion among the highest ramping hospitals suggests that severe congestion can arise even without being among the busiest ambulance destinations. Hospital specific operational characteristics influence ramping beyond patient volume alone.
Finally, some hospitals exhibit relatively stable ramping throughout the day (H01, H22), while others experience pronounced afternoon surges and evening surges (18:00-20:00). These differences indicate considerable temporal heterogeneity across hospitals.
Reordering hospitals according to their case loads in Fig. \ref{fig:hospP90Ramping} ramping provides additional insight into the severity of congestion with respect to the ambulance case volume.
Several hospitals identified as having the most severe ramping (particularly H03-H05, H07, H09, and H14, H15) exhibit persistently high p90 values throughout the day, with a morning and evening peak approaching 144 and 153 minutes respectively. These hospitals appear to experience sustained operational pressure rather than occasional spikes in congestion, with only a brief easing in the early morning around 05:00–06:00. H14, H15 hospitals receive comparatively fewer ambulance cases but nevertheless exhibit substantial ramping throughout the day compared to H06-H08, H10 and H12. 
In contrast, several high volume hospitals shown in Fig. \ref{fig:hospP90Ramping} are absent from Fig. \ref{fig:hospP90RampingOverall} ranking, indicating that high ambulance case load does not necessarily translate into the most severe ramping. This reinforces the observation that factors such as capacity, inpatient bed availability, discharge processes, staffing levels and operational practices also contribute to ambulance offload delays.
\subsection{Hospital Arrival Concurrency and Its Association with Ramping }A recurring question in EMS operations analysis is whether the current state of the receiving hospital, in particular, the number of ambulances already waiting or recently arriving at the emergency department, influences the offload/ramping outcome of the arriving ambulance \cite{li2026empirical, LI2022101146}. Or in other words, beyond long run hospital characteristics and time of day patterns, does the transient hospital congestion state at the moment of arrival matter?
For each case in the ED cohort, the number of prior ambulance arrivals at the same destination hospital in the preceding 60 minutes was computed (1h prior arrivals). Prior arrivals were counted from all ambulance arrivals at the receiving hospital, irrespective of transport priority, since any ambulance competes for the same ED resources. Across 2,026,591 million cases\footnote{Denominator vary marginally (2,026,591) from a ramping cohort (noted earlier in Data Cleaning and Assessment- 2,026,597) where a computation requires grouping on additional fields; differences affect 6 records and no reported statistic. These 6 records from a single incident are collapsed when grouping by case.} with concurrency computed (within this cohort each hospital has its own case count), the mean prior hour arrival count was 2.46 (SD 2.07, range 0-24, median 2).
Ramping outcomes rose monotonically with concurrency. Cases arriving at hospitals with zero prior arrivals in the preceding hour experienced a median ramping of 23.6 minutes and a 90th percentile of 66.1 minutes. Cases arriving at hospitals with 10 or more prior arrivals experienced a median ramping of 45.2 minutes and a 90th percentile of 127.9 minutes which is approximately 1.92 times the corresponding median and p90 values for the zero concurrency cases.

\begin{figure}[!t]
\centering
\includegraphics[width=\columnwidth]{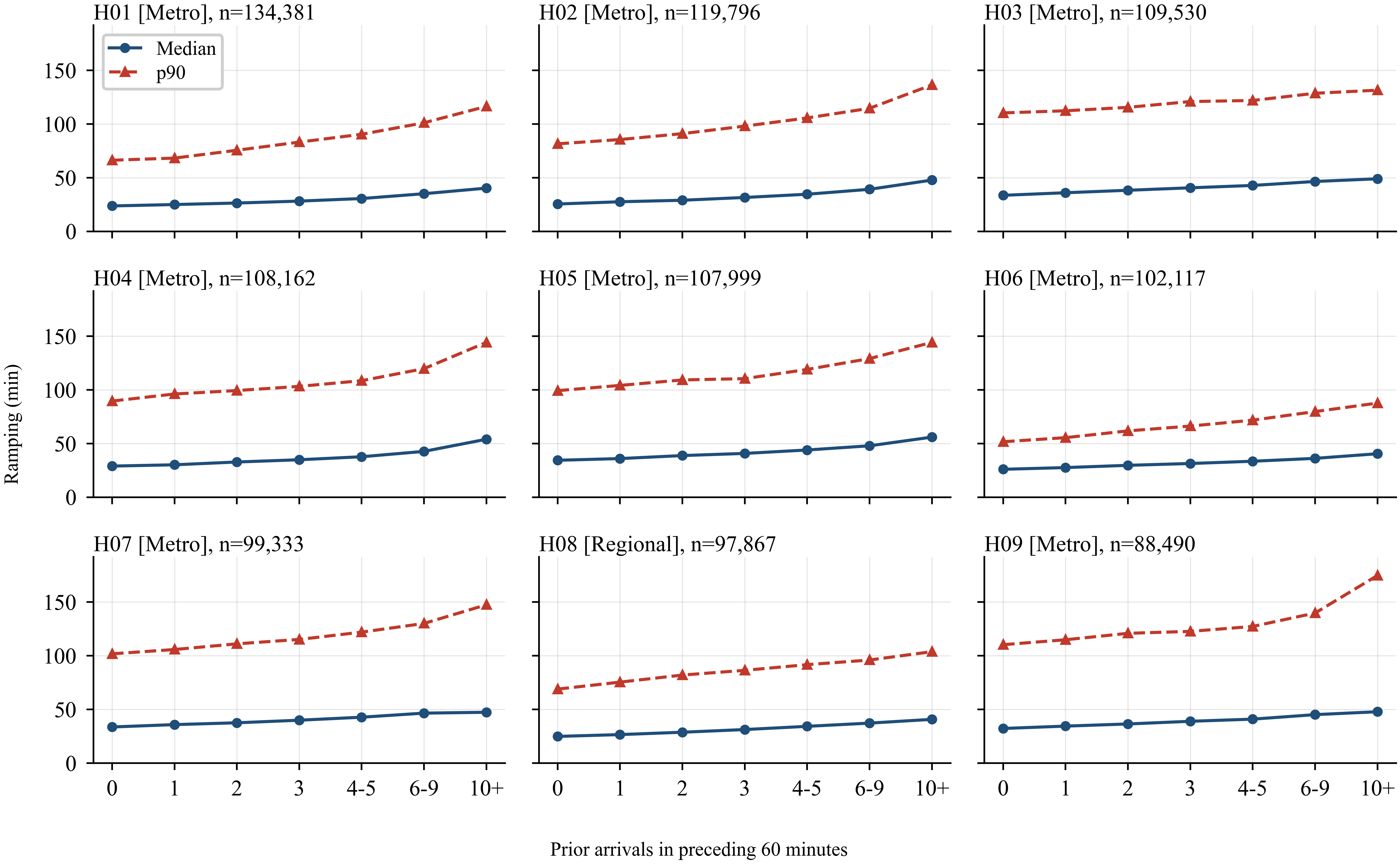}%{figures/concurrency_perHospital_9Hospital.png}
\caption{Median and 90th-percentile ramping durations by concurrency bin for the nine highest–case load ED hospitals. Each panel corresponds to a single hospital. Blue circles indicate median ramping; red triangles indicate the p90 value.}
\label{fig:perHospitalConcurrency}
\end{figure}

\begin{table}[!t]
\centering
\caption{Relationship between prior ambulance arrivals within the preceding hour (hospital concurrency) and ramping duration.}
\label{tab:concurrency_ramping}
\footnotesize
\begin{tabular}{lrrrr}
\toprule
\textbf{Concurrency} & \textbf{Cases} & \textbf{Median} & \textbf{p75} & \textbf{p90} \\
\textbf{bin} & ($n$) & (min) & (min) & (min) \\
\midrule
0    & 388,362 & 23.6 & 38.0 & 66.1 \\
1    & 406,521 & 28.4 & 47.6 & 82.4 \\
2    & 365,989 & 31.6 & 53.7 & 91.8 \\
3    & 302,032 & 34.1 & 58.1 & 97.4 \\
4-5  & 380,616 & 36.6 & 62.4 & 103.5 \\
6-9  & 176,858 & 40.2 & 68.8 & 112.6 \\
10+  & 6,213   & 45.2 & 77.2 & 127.5 \\
\bottomrule
\end{tabular}
\end{table}
Concurrency is mechanically relates to the time of the day, as peak hours generate both more prior arrivals and longer ramping. The unstratified gradient in Table \ref{tab:concurrency_ramping} is therefore cannot by itself be interpreted as evidence of an independent effect of concurrency.
%To test that concurrency is correlated with time of day with peak hours generating both more prior arrivals and typically longer ramping (independent concurrency effect), the unstratified pattern alone is insufficient evidence (Table \ref{tab:concurrency_ramping}). 
To distinguish these two possibilities, ramping was stratified simultaneously by both hour of day and concurrency bin. Fig. \ref{fig:OneHourBeforeArrivalRampingHeatMap} presents the resulting heatmaps: rows correspond to hour-of-day (0 to 23), columns correspond to concurrency bin (0, 1, 2, 3, 4-5, 6-9, 10+), and cell values on the right heatmap are the p90 ramping for that specific hour and concurrency combination whereas left heatmap shows the corresponding median values. The patterns visible can be interpreted as follows: (i) If ramping increases from left to right within a single row (that is, at constant hour of day, higher concurrency yields higher ramping), then concurrency has a real effect that persists after controlling for time. (ii) If ramping increases from top to bottom (or vise versa) within a single column [at constant concurrency, later (or earlier) hours have higher ramping], then time of day matters independently of concurrency. And/or (iii) if both patterns are visible simultaneously, then both effects are real and independent. The heatmap shows both patterns clearly. Ramping generally increases with concurrency across the columns within each hour of the day. Ramping varies across hours of the day within each concurrency bin, with higher values generally observed during the 12:00–20:00 period, except in the 0-concurrency bin. Concurrency and hour of day contribute independently to ramping outcomes. %The heatmaps provide descriptive evidence of associations between both concurrency and hour of day and ramping. 
To verify that this aggregated pattern is not driven by a small number of hospitals in study cohort, the concurrency analysis was repeated at the individual hospital level for the hospitals arranged by their case loads. Fig. \ref{fig:perHospitalConcurrency} presents the results as a 3x3 grid of small multiples, with each panel showing median and p90 ramping across concurrency bins for a single hospital (only 9 hospitals are shown here for image clarity. The table for all the hospitals is available).
%\footnote{Only 9 hospitals are shown here for image clarity. The table for all the hospitals is available.} 
The pattern demonstrates that the positive association between hospital concurrency and ambulance ramping is consistently observed across the hospital with EDs, indicating that the aggregate relationship is not driven by a small subset of hospitals. Median ramping increases progressively with the number of ambulance arrivals during the preceding hour for every hospital, suggesting that typical offload delays worsen as local ambulance arrivals accumulates. The p90 curves reveal greater interhospital variability. At some hospitals (for example, H04, H05, H07 and H09), the longest delays rise sharply as more ambulances arrive. At others, such as H14, the longest delays stay high regardless of how many ambulances have recently arrived, remaining between 127 and 138 minutes across all concurrency levels. These differences imply that hospitals vary not only in their baseline congestion but also in their operational sensitivity to additional ambulance arrivals. Collectively, the results demonstrate that recent ambulance arrival volume is a robust predictor of subsequent ramping across diverse hospitals. They also underscore hospital-specific heterogeneity, supporting the inclusion of both concurrency metrics and hospital identity in predictive models for ambulance routing and congestion prediction. % An overall increasing trend is evident across almost all hospitals, although the magnitude and steepness of the relationship vary between hospitals.
We can infer that concurrency and time of day contribute independently to ramping in the observed data, but these observational analysis does not quantify the causal reduction in ramping that may result from actively redistributing \cite{Ong2025AmbulanceDiversion} ambulance arrivals across hospitals. Causal claims require intervention studies or, at minimum, natural experiment analyses of specific policy changes affecting concurrency without affecting the confounders that co-vary with hospital selection. Such extensions are beyond the scope of the current analysis.
\section{Conclusion}
The analysis began with a system that appears stable in aggregate where annual case loads varied little across the study window. Beneath that stability, however, the same caseload consumed progressively more ambulance capacity, with cumulative AHL reaching 1,491,127 hours and annual losses rising 57\% to their 2022 peak while ED-bound transports fell by 3.7\% over the same period (Table \ref{tab:yearly_transport_arrival_priority_volume}). Tracing \emph{where} that capacity is lost revealed a bounded concentration in 25 to 30 hospitals; tracing \emph{when} revealed a 2 to 4 hour propagation from demand surge to handover delay; and tracing under \emph{what} conditions revealed an effect of recent arrival volume at the receiving hospital where the number of arrivals in the preceding hour exerts an effect on handover duration that persists within hour of the day.
%Tracing \emph{where} geographically revealed two disjoint operational profiles that administrative boundaries do not separate. 
The escalation is therefore driven not by growth in demand but by deterioration in the ambulance to ED interface \cite{yoon2026association}, a deterioration that is concentrated, temporally predictable, and state dependent. Each of these properties constitutes a necessary precondition for prediction, and collectively they delineate the modelling framework that underpins further analysis.
%\subsection{Interpretation of key findings}
\subsection{Identified systemic challenges } %in the Victorian EMS context
The exploratory analysis identifies the following systemic challenges with direct operational significance. First, the burden of ramping is heavily concentrated rather than distributed: the first 10 of 59 ED capable hospitals account for 57.8\% of total Ambulance Hours Lost from 50.9\% of cases, and Metropolitan Melbourne carries 76.7\% of the AHL burden. 
%This concentration is bounded, disproportionality dissipates beyond the top 15 hospitals, implying that system-wide interventions would misallocate effort that targeted intervention at a small set of facilities could capture. 
Second, the system deteriorated substantially within the study window and has not recovered to baseline: annual AHL rose to 57\% from 2020 to its 2022 peak while annual ED bound transports fell slightly, indicating that the escalation is not driven by demand size.
Third, this deterioration coincided with a sustained compositional shift toward higher acuity, the share of transports assigned the most urgent priority code rose from 10\% in 2020 to 13\% in 2023.
%with Code 1 arrivals rising from 42.5\% to 53.4\% of annual volume.
%the system is transporting proportionally sicker patients into progressively slower handover conditions. 
%Fourth, high volume metropolitan LGAs and high per case rural LGAs form disjoint operational profiles with no overlap between their extreme rankings, indicating that a single statewide resourcing logic cannot simultaneously address both. 
Finally, ramping is a state-dependent phenomenon. Hourly ambulance demand is moderately associated with handover delay two to four hours later, and hospital arrival concurrency exerts an effect on ramping that persists within every hour of the day. Together these establish that information relevant to anticipating emerging handover congestion is observable in real time at the point of dispatch, the empirical precondition for hospital-state aware routing.
\subsection{Limitations of this study }Several limitations warrant mention. First, the analysis is confined to the ambulance side operational record: patients who self presented to hospital are not captured, and no ED side data (bed occupancy, staffing, non ambulance inflow, or in-hospital interventions) were available. Ramping is therefore characterised from the ambulance perspective only, and the hospital-side mechanisms underlying the observed patterns, including whether delay is concentrated before or after triage, cannot be resolved from this dataset. Second, LGAs are administrative rather than operational units. They are used here to attribute cases to the communities in which incidents occurred, but a single LGA may encompass areas differing substantially in remoteness, population density and distance to the nearest emergency department, and LGA level statistics therefore average over heterogeneous conditions. The analysis of access and equity that LGA boundaries would support, such as, comparing service levels across communities with differing socioeconomic or demographic profiles, is beyond the scope of the present study but represents a natural extension. Third, the study is observational and exploratory by design; the reported associations, including the concurrency–ramping gradient, do not establish that intervening on these factors would proportionally reduce ramping. Fourth, several analytical choices bound the findings: ramping durations above 600 minutes were excluded from the analytical cohort and a same cap was applied in AHL summation; demand was defined at the dispatch level, with multi unit responses contributing one record per responding unit; and selected analyses were stratified to non-time critical, non-urgent transports also. Fifth, the study window begins in January 2020 and therefore contains no pre-pandemic baseline. Victoria’s 2020 lockdowns affected ambulance demand and hospital activity \cite{1c0d44cf377646259d12e84bff7b56ac}; easing restrictions later restored demand amid ongoing pandemic operational and infection-control requirements. Consequently, the longer handover and ramping delays observed in 2021–2022 likely reflect the unusual mix of rebounding demand and pandemic-era operating conditions, not a broader decline in system performance. Similarly, the lower delays in 2023–24 cannot, from these data, be taken as evidence of improved ramping performance. Without pre-2020 observations, the study cannot distinguish genuine improvement in the later years from post-pandemic normalisation. Finally, the findings derive from a single statewide jurisdiction with a specific hospital geography and dispatch system; the analytical framework is transferable, but the specific empirical patterns may not generalise to other EMS systems. 
%Without pre-2020 observations representing more typical operating conditions, the study cannot establish whether the later years represent genuine improvement, post-pandemic normalisation, or both. 
%Fifth, the study window begins in January 2020 and therefore contains no pre-pandemic baseline. Victoria's lockdowns affected both ambulance demand and hospital occupancy during 2020, so handover/ramping delays in that year is unlikely to reflect normal operating conditions. This affects the results as they reflect recovery from an atypically year. 
%Second, there were limited data available regarding in-hospital interventions. Finally, a significant pro-portion of patients were not successfully linked to the VAED and VEMD databases and were therefore not included in the primary analysis. This could represent a source of selection bias.

\appendices

\section*{Acknowledgment} This work is supported by the Next Gen Graduate Program, a
collaborative program between the Commonwealth Scientific and Industrial Research Organisation (CSIRO), Australia and Victoria University (The Institute for Sustainable Industries and Liveable Cities, The Centre of Excellence in Paramedicine), with Ambulance Victoria as an industry partner. The data for this study have been obtained from the Research Governance, Centre for Research \& Evaluation, Ambulance Victoria. 
%The authors would like to thank AV's Research Governance, Centre for Research \& Evaluation for providing the data set.
\ifCLASSOPTIONcaptionsoff
  \newpage
\fi
\bibliographystyle{IEEEtran}
\bibliography{bibliography}
\end{document}